\documentclass[10pt,twocolumn,letterpaper]{article}

\usepackage[pagenumbers]{wacv} 

\definecolor{rankfirst}{RGB}{192,0,0}
\definecolor{ranksecond}{RGB}{230,145,56}
\definecolor{rankthird}{RGB}{47,84,41}

\newcommand{\method}{GLOW-FM}

\usepackage{algorithm}
\usepackage{algpseudocode}
\usepackage{tikz}
\usepackage{multirow}
\usepackage{colortbl}

\definecolor{wacvblue}{rgb}{0.21,0.49,0.74}
\usepackage[pagebackref,breaklinks,colorlinks,allcolors=wacvblue]{hyperref}

\usepackage{amsthm}
\newtheorem{lemma}{Lemma}
\newtheorem{theorem}{Theorem}
\usepackage{siunitx}

\newcommand{\mstdsci}[2]{%
  #1{\scriptsize\(\mathord{\pm}\num[
    exponent-mode=scientific,
    exponent-product=\mathord{\times},
    round-mode=figures,
    round-precision=1,
    round-pad=true
  ]{#2}\)}%
}

\def\confName{WACV}
\def\confYear{2027}

\title{GLOW-FM: Efficient Generative Low-Light Image Enhancement with One-Step \\ Latent Flow Matching}

\author{
Ariel Lapid\thanks{Both authors contributed equally to this work}
\quad
Yuval Haitman\footnotemark[1]
\quad
Reuven Peretz
\quad
Arnon Netzer
\quad
Idit Diamant\thanks{Corresponding author:  
\texttt{Idit.Diamant@arm.com}} \\
Arm AI Research (AAIR), Arm Ltd
}

\usepackage[subtle]{savetrees}

\begin{document}
\maketitle
\begin{abstract}
Low-light image enhancement (LLIE) aims to recover well-exposed and perceptually faithful images from severely degraded observations. Recent diffusion-based generative methods achieve strong perceptual quality by leveraging powerful image priors, but their iterative sampling incurs substantial computational cost, limiting practical deployment. Despite recent efficiency gains in generative LLIE, high-quality enhancement with a single function evaluation and low overall computation remains challenging, particularly under diverse real-world low-light conditions.
We introduce \textbf{\method{}}, a one-step latent flow-matching framework for efficient generative LLIE. \method{} performs conditional transport in the latent space of a frozen Tiny-VAE and leverages Terminal Velocity Matching (TVM) to enable enhancement with a single function evaluation. Matched comparisons with alternative one-step objectives support TVM as an effective design choice, and, to the best of our knowledge, this is its first adaptation to image restoration and specifically to LLIE.
Extensive experiments on challenging real-world benchmarks demonstrate that \method{} achieves competitive image quality and state-of-the-art cross-dataset generalization while requiring only a single function evaluation and 34 GFLOPs for the small variant. These findings highlight the potential of \method{} to bring generative LLIE closer to practical deployment through single-step inference and compact latent modeling.
Code is available at \url{https://github.com/arm-research/AAIR-GLOW-FM}.
\end{abstract}

\vspace{-10pt}

\section{Introduction}
\label{sec:intro}

\begin{figure}[t]
    \centering
    \includegraphics[width=\linewidth, trim=0cm 1.5cm 0cm 0.5cm,
  clip]{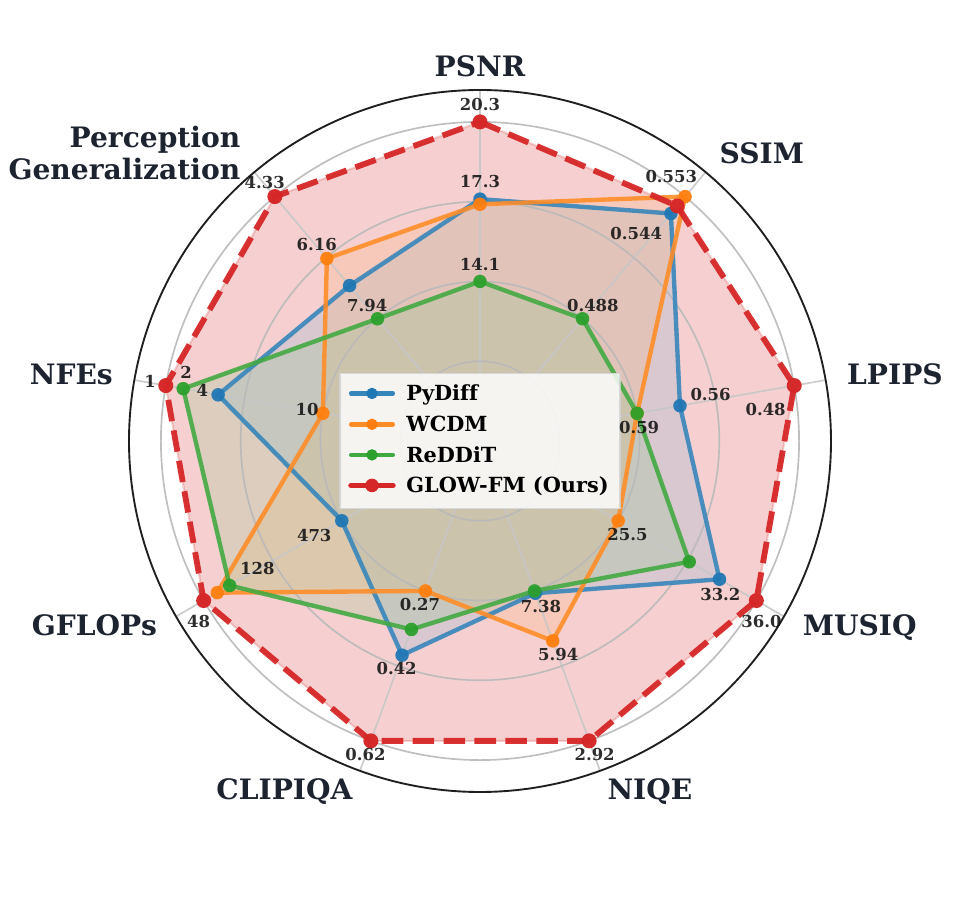}
\caption{\textbf{\method{} Performance.}
Comparison with efficient generative LLIE baselines on LSD-DLL.
Metrics are min--max normalized across the compared methods, with
lower-is-better metrics reversed so that farther from the center is
always better; axis annotations show the corresponding absolute metric
values. Full results are reported in \cref{tab:paired_lsd_uhd}, while
\textit{Perception Generalization} uses LSD-U NIQE from
\cref{tab:lsd_u}.}
\label{fig:dll-radar}\vspace{2pt}
\end{figure}

Low-light image enhancement (LLIE) aims to recover well-exposed and visually pleasing images from observations captured under insufficient illumination. Real-world low-light images often exhibit a complex combination of degradations, including severe underexposure, sensor noise, color distortion, reduced contrast, detail loss, and local saturation, which vary significantly across cameras and imaging conditions, making LLIE a challenging problem in real-world imaging~\cite{Ciubotariu_2026_CVPRW,chen2018learning,li2021low}. Beyond degrading visual quality, these artifacts also impair downstream computer vision tasks such as object detection and recognition, whose performance declines substantially on poorly exposed inputs~\cite{loh2019getting}. Consequently, LLIE seeks not only to enhance image appearance but also to restore perceptually faithful content that supports reliable high-level vision applications~\cite{FeatEnHancer_Hashmi_ICCV23,MorawskiCLH21}.

Regression-based LLIE methods provide strong distortion performance and efficient inference~\cite{multinex2026,cai2023retinexformer,yan2025hvi,zhang2021beyond,snr_aware,zhang2022deep,cpga_net,lolv1,uretinexnet,lsd_dataset,uhd_lol_4k_dataset}. These approaches typically optimize distortion-oriented objectives and are effective at recovering visibility and suppressing noise, but often struggle under severe degradation, where substantial information loss makes enhancement inherently ambiguous. Trained with pixel-wise objectives, they tend to produce overly smooth results and suppress fine details~\cite{blau2018perception}.
To address this limitation, generative models~\cite{gsad,diffretinex,Jiang_2024_ECCV,wang2023exposurediffusion,yin2023cle,jiang2021enlightengan,he2025reti} have emerged as a powerful LLIE paradigm, learning the distribution of natural images and leveraging strong image priors to recover realistic textures and structures that are difficult to reconstruct with deterministic networks. These methods demonstrate substantial improvements in perceptual quality and robustness under severe degradations.
However, these gains come at the cost of iterative sampling, resulting in high computational complexity and limiting practical deployment, particularly for high-resolution enhancement in resource-constrained applications such as smartphone camera image processing.

To address this limitation, several efficient generative LLIE frameworks have recently been proposed~\cite{wcdm, zhou2023pydiff, reddit}. These methods significantly reduce the required number of function evaluations (NFEs) compared to conventional diffusion models. Nevertheless, they still rely on iterative refinement procedures and diffusion-inspired architectures whose computational complexity remains substantial. For example, WCDM~\cite{wcdm} requires 10 NFEs and approximately 90 GFLOPs during inference, while ReDDiT~\cite{reddit} reduces sampling to 2 NFEs but still requires roughly 128 GFLOPs.
Consequently, while recent generative LLIE approaches have substantially reduced sampling cost, achieving high-quality enhancement with a single inference step and a low overall computational budget remains an active challenge, particularly across diverse and demanding real-world low-light conditions.

In this work, we address this challenge by introducing an efficient \textbf{G}enerative \textbf{Low}-Light image enhancement with one-step latent \textbf{F}low \textbf{M}atching (\textbf{\method{}}).
Our approach leverages Terminal Velocity Matching (TVM)~\cite{tvm}, which enables direct prediction of the normal-light image, eliminating the need for iterative sampling during inference.
Comparisons with standard Flow Matching~\cite{lipman2023flow,rectifiedflow}, Consistency Flow Matching~\cite{cfm}, and MeanFlow~\cite{meanflow}
establish Terminal Velocity Matching (TVM)~\cite{tvm} as a deliberate design choice for single-step LLIE. To the best of our knowledge, this is the first adaptation of TVM to image restoration, and specifically to LLIE.
\method{} learns a conditional transport that maps degraded low-light representations directly to their well-exposed counterparts in the latent space of a frozen Tiny-VAE~\cite{taesd}. Operating in this compact latent space substantially reduces computational cost while preserving the expressive power of generative modeling. Moreover, latent representations capture semantic image structure while suppressing dataset-specific variations, improving restoration and generalization~\cite{oldphotorestoration,ldrps,thomas2025latent}. 

Extensive experiments on challenging real-world low-light benchmarks~\cite{lsd_dataset, uhd_lol_4k_dataset} demonstrate that \method{} provides an effective generative prior for low-light enhancement while operating under small inference budgets. Despite requiring only a single NFE, our method achieves image quality competitive with recent efficient generative LLIE approaches and consistently exhibits strong cross-dataset generalization as shown in \cref{fig:dll-radar}. 
These results suggest that latent flow matching, combined with one-step TVM estimation, offers a promising direction toward practical, high-quality, and computationally efficient low-light image enhancement.

\noindent{Our main contributions are summarized as follows:}
\begin{itemize}
\item We introduce \method{}, an efficient generative framework for low-light image enhancement (LLIE) based on latent flow matching. By performing enhancement in a compact latent space with a lightweight Tiny-VAE and a one-step flow formulation, \method{} achieves high restoration quality at low computational cost.

\item We establish Terminal Velocity Matching (TVM) as the strongest overall one-step objective for LLIE through matched comparisons with Rectified Flow, Consistency Flow Matching, and MeanFlow. To the best of our knowledge, this is the first adaptation of TVM to image restoration, specifically LLIE.

\item We achieve state-of-the-art cross-dataset generalization compared to generative and non-generative baselines, while maintaining competitive image quality and improved efficiency over existing generative LLIE methods.
\end{itemize}

\section{Related Work}
\label{sec:related}
\begin{figure*}[t]
  \centering
  \includegraphics[width=0.95\textwidth, trim=0cm 2.5cm 0cm 0cm,
  clip]{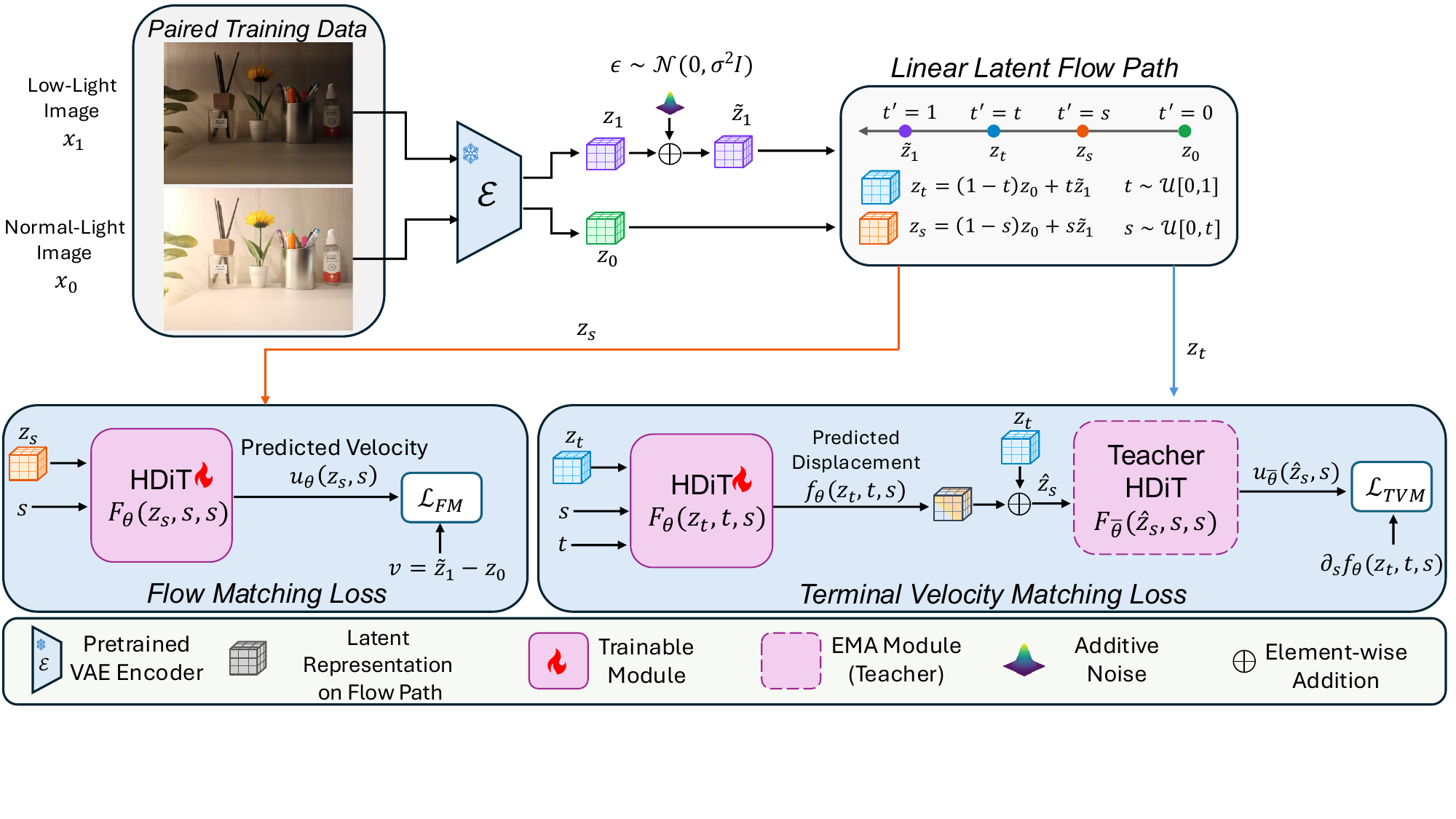}
  \caption{\textbf{Overview of \method{} Training.} Paired low-light and normally exposed images are encoded into a frozen autoencoder latent space. During training, the model learns both the instantaneous flow-matching velocity and a terminal-velocity displacement map between arbitrary times. During inference, the enhanced latent is obtained with a single step and decoded back to image space.}
  \label{fig:method_overview}\vspace{-10pt}
\end{figure*}
\subsection{Generative Low-Light Image Enhancement}
A large body of LLIE research focuses on directly regressing a normal-light image from a low-light input using Retinex decomposition \cite{land1977retinex}, illumination estimation, and deep neural networks~\cite{cai2023retinexformer, yan2025hvi, lsd_dataset, multinex2026}. While these approaches are computationally efficient, they often produce overly smooth results and struggle to capture the diversity of plausible normal-light reconstructions.
The recent success of diffusion models~\cite{ho2020denoising} in image generation has motivated their adoption for high-quality low-light image enhancement~\cite{diffretinex,gsad,reddit,zhou2023pydiff,wcdm, he2025reti,Jiang_2024_ECCV}.
Diff-Retinex~\cite{diffretinex} formulates enhancement as Retinex decomposition followed by conditional diffusion-based reconstruction of reflectance and illumination components. GSAD~\cite{gsad} introduces a global structure-aware diffusion framework that combines curvature-based regularization with uncertainty-guided optimization to preserve scene structure and focus restoration on challenging regions. Subsequent work has largely aimed to reduce the high computational cost of diffusion sampling. ReDDiT~\cite{reddit} uses trajectory matching distillation in a reflectance-aware residual space to reduce the NFEs required at inference. PyDiff~\cite{zhou2023pydiff} performs denoising in a coarse-to-fine resolution pyramid and employs a global correction module to mitigate brightness and color inconsistencies. WCDM~\cite{wcdm} accelerates inference by conducting diffusion in the wavelet domain on low-frequency sub-bands while restoring high-frequency details through a dedicated branch. 
Reti-Diff~\cite{he2025reti} generates compact latent Retinex priors
to guide a transformer-based restoration network, while
LightenDiffusion~\cite{Jiang_2024_ECCV} combines latent Retinex
decomposition with diffusion for unsupervised LLIE. Related concurrent
preprints, IllumFlow~\cite{illumflow} and
Consist-Retinex~\cite{consistretinex}, further explore one-step
generative LLIE through Retinex-based decomposition.

\subsection{Generative Model Acceleration}\vspace{-5pt}
Generative diffusion and flow models have achieved remarkable success
across a wide range of image-generation tasks. Their strong priors have
also motivated applications to image restoration, where the goal is to
recover a clean image from a degraded observation. InDI~\cite{indi}
performs restoration through direct iteration, while
IR-SDE~\cite{irsde} models the restoration trajectory using a
mean-reverting stochastic process. Bridge-matching
methods~\cite{dsbm} further formalize transport between degraded and
clean endpoint distributions, and flow-based approaches such as
FlowIE~\cite{flowie} and PnP-Flow~\cite{pnpflow} extend this formulation
to image restoration.
Despite their strong restoration capabilities, these methods commonly
require iterative sampling, resulting in substantial inference cost.
Early acceleration approaches employed fast post-hoc
samplers~\cite{ddim,dpmsolver,edm}, reducing inference from hundreds of
steps to tens, often at the expense of quality under aggressive
acceleration.
More recently, flow-based objectives have been developed specifically
for efficient one- and few-step generation. Rectified
Flow~\cite{rectifiedflow} learns straighter transport trajectories,
while Consistency Flow Matching~\cite{cfm} regularizes the velocity
field for few-step inference. ELIR~\cite{elir} applies Consistency Flow
Matching to image restoration in the latent space of a Tiny
AutoEncoder. MeanFlow~\cite{meanflow} learns mean transport trajectories,
whereas TVM~\cite{tvm} learns finite displacements between arbitrary
timesteps for high-fidelity one- and few-step generation.
Despite these advances, efficient flow-based generation remains
relatively unexplored for LLIE. Building upon TVM, we introduce a
single-step latent flow-matching framework for efficient low-light
enhancement.

\section{Method}
\label{sec:method}

We formulate low-light image enhancement as a conditional flow-matching problem~\cite{lipman2023flow} between the low-light
distribution $p_1(x_1)$ and the normally exposed distribution $p_0(x_0)$. Given a paired sample $(x_1,x_0)$, where $x_1\sim p_1(x_1)$ and $x_0\sim p_0(x_0)$ denote the low-light source and normal-light target, respectively, we follow the TVM time convention~\cite{tvm}: the source is placed at $t=1$ and the target at $t=0$, opposite to the original Flow Matching convention. Enhancement therefore proceeds toward earlier terminal times, from $t=1$ to $t=0$.

Rather than learning this transport directly in pixel space, we perform it in the latent space of a frozen VAE with encoder $\mathcal{E}$ and decoder $\mathcal{D}$. We denote the corresponding latents as $z_1=\mathcal{E}(x_1)$ and $z_0=\mathcal{E}(x_0)$, and define intermediate states $z_t$ along the flow-path. To enable efficient inference, we train a two-time conditioned model $F_\theta(z_t,t,s)$ with Terminal Velocity Matching (TVM)~\cite{tvm},  which learns displacements along the trajectory rather than relying only on numerical integration of an instantaneous velocity field. This allows the same latent flow model to support both one-step enhancement and few-step refinement. Beyond its reported strong one-step performance, we motivate TVM for restoration through theoretical analysis. Under the TVM regularity conditions~\cite[Theorem~1]{tvm} and a Lipschitz assumption on the decoder, we derive a bound on the pixel-space Wasserstein-2 distance~\cite{panaretos2019statistical} between the restored-image and target normal-light distributions in terms of the TVM objectives and autoencoder reconstruction error. This links TVM training to pixel-space restoration; the complete theorem and proof are provided in \cref{app:tvm_w2_bound}. We next describe TVM’s adaptation to our restoration setting, training and inference, and the design choices enabling efficient generation.

\subsection{Terminal Velocity Matching}
\label{subsec:tvm}

Flow Matching~\cite{lipman2023flow} learns an instantaneous velocity field and generates samples by numerically integrating this field over time. To accelerate this process, TVM directly parameterizes the finite transition between two distinct time steps, bypassing the need for small integration steps.
For an ideal latent velocity field $u(z_t,t)$, the flow map from time $t$ to an earlier time $s\leq t$ is given by:\vspace{-5pt}
\begin{equation}
  z_s := \psi(z_t,t,s)=z_t+\int_t^s u(z_r,r)\,dr .\vspace{-6pt}
\end{equation}
The corresponding net displacement integrates the velocity from $t$ to $s$, while its derivative at the zero-displacement boundary recovers the instantaneous velocity:
\begin{equation}
\scalebox{0.92}{$
  f(z_t,t,s)=\int_t^s u(z_r,r)\,dr,\quad \partial_s f(z_t,t,s)|_{s=t}=u(z_t,t).
$}
\end{equation}
Following TVM, we parameterize the displacement with a two-time network. A scaling is used to enforce zero displacement at the boundary $s=t$, where the instantaneous velocity is also recovered:\vspace{-5pt}
\begin{equation}\label{eq:tvm_param}
\begin{aligned}
  &f_\theta(z_t,t,s) = (s-t)F_\theta(z_t,t,s),\\
  &u_\theta(z_t,t) := \left.\partial_s f_\theta(z_t,t,s)\right|_{s=t}
  = F_\theta(z_t,t,t).
\end{aligned}\vspace{-3pt}
\end{equation}
The boundary relation above recovers standard flow matching when the displacement length goes to zero. For finite-step inference, however, the desired objective is the full displacement from time $t$ to an earlier time $s$. A direct objective would regress this displacement by:\vspace{-7pt}
\begin{equation}
  \mathcal{L}_{\mathrm{displ}}
  =
  \mathbb{E}
  \left[
  \left\|
  f_\theta(z_t,t,s)
  -
  \int_t^s u(z_r,r)\,dr
  \right\|_2^2
  \right],\vspace{-3pt}
\end{equation}
but this target is unavailable because it requires the true velocity field along the entire trajectory. TVM avoids this explicit integration by using the terminal velocity, defined as the derivative of the displacement with respect to the terminal time $s$. In the ideal case, this derivative matches the velocity at the terminal point, i.e., $\partial_s f(z_t,t,s)=u(z_s,s)$ for any $s\leq t$.

Since the true terminal point $z_s$ and velocity field are unavailable for the predicted displacement, we use the predicted terminal latent \vspace{-5pt}
\begin{equation}\label{eq:predicted_zs}
  \hat{z}_s = z_t + f_\theta(z_t,t,s),
\end{equation}
and approximate the terminal velocity with the model velocity,
\begin{equation}
  \partial_sf_\theta(z_t,t,s)
  \approx
  u_\theta(\hat{z}_s,s)
  =
  F_\theta(\hat{z}_s,s,s).
\end{equation}
This provides a practical proxy for supervising finite displacements without explicit ODE integration, and it is the basis of the TVM loss described in the following section.
\begin{figure*}[t]
  \centering
  \includegraphics[width=0.9\textwidth]{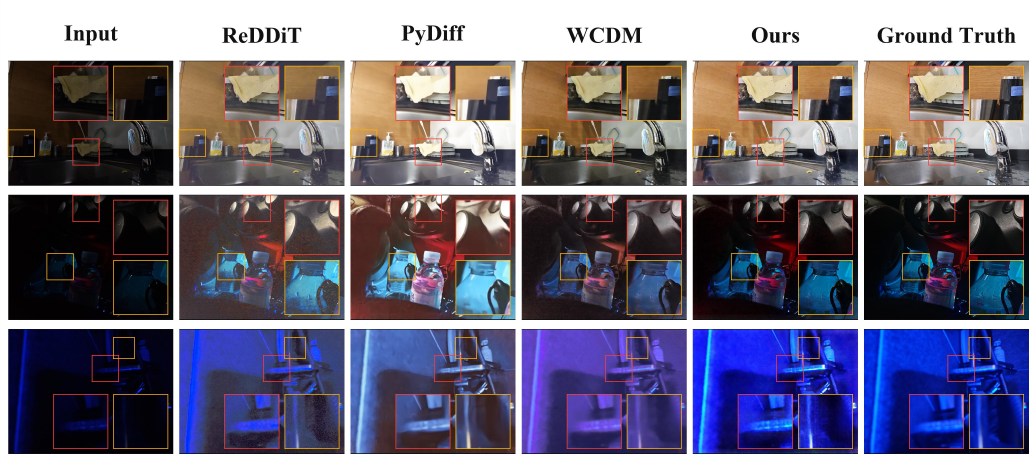}
  \caption{\textbf{Qualitative comparison on LSD-DLL.} Representative examples from LSD-DLL comparing low-light inputs, outputs from ReDDiT, PyDiff, WCDM, and \method{}, and the corresponding ground-truth normal-light images.}\vspace{-10pt}
  \label{fig:lsd_dll_qualitative}
\end{figure*}
\subsection{Training}
\label{subsec:training}
\begin{algorithm}[t]
  \caption{Training \method{}}
  \label{alg:training}
  \begin{algorithmic}[1]
    \Require Paired images $(x_1,x_0)$, noise scale $\sigma$, encoder $\mathcal{E}$, weights $\theta$, EMA weights $\bar{\theta}$
    \For{training step $\tau$}
      \State Encode $z_1=\mathcal{E}(x_1)$ and $z_0=\mathcal{E}(x_0)$
      \State Sample $\epsilon\sim\mathcal{N}(0,I)$ and set $\tilde{z}_1=z_1+\sigma\epsilon$
      \State Sample $t\sim\mathcal{U}[0,1]$ and $s\sim\mathcal{U}[0,t]$
      \State Form $z_t=(1-t)z_0+t\tilde{z}_1$ and $z_s=(1-s)z_0+s\tilde{z}_1$
      \State Set target velocity $v=\tilde{z}_1-z_0$
      \State Compute $\mathcal{L}_{\mathrm{FM}}$ using \cref{eq:fm_loss}
      \State Compute $\hat{z}_s=z_t+f_\theta(z_t,t,s)$
      \State Compute target proxy $u_{\bar{\theta}}(\hat{z}_s,s)=F_{\bar{\theta}}(\mathrm{sg}(\hat{z}_s),s,s)$
      \State Compute $\mathcal{L}_{\mathrm{TVM}}$ using \cref{eq:tvm_loss}
      \State Update $\theta$ with $\mathcal{L}_{\mathrm{FM}}+\lambda\mathcal{L}_{\mathrm{TVM}}$
      \State Update EMA parameters $\bar{\theta}$
    \EndFor
  \end{algorithmic}
\end{algorithm}\vspace{-5pt}
\begin{algorithm}[t]
  \caption{Inference with \method{}}
  \label{alg:inference}
  \begin{algorithmic}[1]
    \Require Low-light image $x_1$, $\mathcal{E},\mathcal{D}$, weights $\theta$, NFEs $M$
    \State Initialize $z^{(M)}\!=\!\mathcal{E}(x_1)$; choose grid $1\!=\!t_M\!\!>\!\!\cdots\!\!>\!t_0\!\!=\!0$
    \For{$m=M,M-1,\ldots,1$}
      \State $z^{(m-1)} = z^{(m)} + f_\theta(z^{(m)}, t_m,t_{m-1})$
    \EndFor
    \State Decode $\hat{x}_0=\mathcal{D}(z^{(0)})$
    \State \Return enhanced image $\hat{x}_0$
  \end{algorithmic}
\end{algorithm}
\cref{fig:method_overview} illustrates the training procedure, and \cref{alg:training} provides the detailed algorithmic steps. Given paired low-light and normally exposed images $(x_1,x_0)$, we first encode them via the frozen VAE encoder as $z_1=\mathcal{E}(x_1)$ and $z_0=\mathcal{E}(x_0)$. Directly using the low-light latent endpoint can make the source distribution overly concentrated around the observed degraded embeddings, which may lead the model to learn a sharp and less stable transport. To smooth the source distribution and make it better supported over the target latent space, we follow stochastic-interpolant practice~\cite{albergobuilding,elir} and perturb the source latent with small Gaussian noise:\vspace{-6pt}
\begin{equation}
  \tilde{z}_1 = z_1 + \sigma \epsilon,
  \qquad
  \epsilon \sim \mathcal{N}(0,I).\vspace{-5pt}
\end{equation}
The perturbed latent $\tilde{z}_1$ is then used as the source endpoint during training.

\vspace{3pt}
\noindent\textbf{Flow-matching loss.}
We train on the linear latent flow path between the normally exposed target $z_0$ and the perturbed low-light source $\tilde{z}_1$. For sampled times $t\!\!\sim\!\!\mathcal{U}[0,1]$ and $s\!\!\sim\!\!\mathcal{U}[0,t]$, we define \vspace{-7pt}
\begin{equation}
  z_t=(1-t)z_0+t\tilde{z}_1,
  \qquad
  z_s=(1-s)z_0+s\tilde{z}_1,\vspace{-5pt}
\end{equation}
with the conditional flow-matching target: $v = \tilde{z}_1-z_0$.
The FM loss supervises the boundary case of the two-time model, where $s\!=\!t$ and $F_\theta$ represents the instantaneous velocity (\cref{eq:tvm_param}):
\begin{equation}\label{eq:fm_loss}
  \mathcal{L}_{\mathrm{FM}}
  =
  \mathbb{E}_{z_0, z_1, s,\epsilon}
  \left[
  \left\|
  F_\theta(z_s,s,s)-v
  \right\|_2^2
  \right].\vspace{-3pt}
\end{equation}
This anchors the model to the standard flow-matching velocity and provides the velocity proxy used by TVM at terminal points.

\vspace{3pt}
\noindent\textbf{Terminal-velocity loss.}
The TVM loss supervises finite displacements between two sampled times. For each pair $(s,t)$ such that $s\leq t$, the displacement model $f_\theta(z_t,t,s)$ predicts a terminal latent $\hat{z}_s$ as defined in \cref{eq:predicted_zs}. Following TVM~\cite{tvm}, we train the displacement by matching its derivative with respect to the terminal time $s$ to the velocity at the predicted terminal point. Using the parameterization in \cref{eq:tvm_param}, this couples the derivative of the finite-displacement prediction, $\partial_s f_\theta(z_t,t,s)$,  with the boundary velocity branch $F_\theta(\hat{z}_s,s,s)$. Since the true terminal velocity is unavailable, we use an EMA copy of the network as a target proxy:\vspace{-5pt}
\begin{equation}
  u_{\bar{\theta}}(\hat{z}_s,s)
  =
  F_{\bar{\theta}}
  \left(
  \mathrm{sg}(\hat{z}_s),s,s
  \right),\vspace{-3pt}
\end{equation}
where $\bar{\theta}$ denotes the EMA parameters (teacher model) and $\mathrm{sg}(\cdot)$ denotes stop-gradient. The terminal-velocity loss is
\begin{equation}\label{eq:tvm_loss}
  \mathcal{L}_{\mathrm{TVM}}
  =
  \mathbb{E}_{z_0,z_1,t,s,\epsilon}
  \left[
  \left\|
  \partial_s f_\theta(z_t,t,s)
  -
  \mathrm{sg}\!\left(u_{\bar{\theta}}(\hat{z}_s,s)\right)
  \right\|_2^2
  \right].\vspace{-5pt}
\end{equation}
The derivative $\partial_s f_\theta(z_t,t,s)$ is computed with a Jacobian-vector product (JVP) with respect to the terminal time $s$. The EMA target and stop-gradient follow the practical biased estimator used in TVM: the target velocity is treated as fixed during the current update, which prevents gradients from flowing through both the predicted terminal point and the target branch. This stabilizes the proxy velocity used to supervise finite displacements.

\vspace{3pt}
\noindent\textbf{Total objective.}
The final objective combines the flow-matching and TVM losses from \cref{eq:fm_loss,eq:tvm_loss}:\vspace{-5pt}
\begin{equation}
  \mathcal{L}
  =
  \mathcal{L}_{\mathrm{FM}}
  +
  \lambda
  \mathcal{L}_{\mathrm{TVM}},\vspace{-5pt}
\end{equation}
where $\lambda$ controls the relative contribution of the TVM term. In practice, we employ $\lambda$ as a warm-up schedule, initially training the model with the FM objective to establish a stable velocity field before gradually introducing the TVM loss. 
Additional schedule details are provided in \cref{sec:app_implementation_details}.

\subsection{Inference}
\label{subsec:inference}

At inference time, we encode the low-light input as the initial latent
$z^{(M)}=\mathcal{E}(x_1)$ at the source endpoint $t_M=1$. We use the
uniform time grid $t_m=m/M$, $m=0,\ldots,M$, traversed in decreasing
order from $t_M=1$ to $t_0=0$. For each $m$, the interval
uses start time $t=t_m$ and terminal time $s=t_{m-1}$, and the model
predicts the displacement from the current latent at $t_m$ to the next
latent at $t_{m-1}$. Each interval requires one network evaluation;
hence, the number of intervals $M$ equals the number of NFEs. For
one-step enhancement, $M=1$, so $t=1$ and $s=0$, and the model directly
transports the encoded low-light latent to the enhanced endpoint. For
few-step enhancement, the same displacement model is applied over
multiple shorter intervals. The full inference procedure is summarized
in \cref{alg:inference}.

\begin{table*}[t]
  \centering
  \small
  \setlength{\tabcolsep}{2.5pt}
  \resizebox{\textwidth}{!}{%
\begin{tabular}{@{}l@{\hspace{10pt}}lccccccccccccc@{}}
    \toprule
    \multirow{2}{*}{Category} &
    \multirow{2}{*}{Method} &
    \multirow{2}{*}{NFEs$\downarrow$} &
    \multirow{2}{*}{GFLOPs$\downarrow$} &
    \multirow{2}{*}{Params (M)$\downarrow$} &
    \multicolumn{5}{c}{\textbf{LSD-DLL}} &
    \multicolumn{5}{c}{\textbf{UHD-LOL4K}} \\
    \cmidrule(lr){6-10}\cmidrule(l){11-15}
    & & & & &
    PSNR$\uparrow$ & SSIM$\uparrow$ & LPIPS$\downarrow$ &
    MUSIQ$\uparrow$ & NIQE$\downarrow$ &
    PSNR$\uparrow$ & SSIM$\uparrow$ & LPIPS$\downarrow$ &
    MUSIQ$\uparrow$ & NIQE$\downarrow$ \\
    \midrule

    \multirow{4}{*}{Non-Generative}
    & TFFormer~\cite{lsd_dataset}
    & 1 & 34 & 5.9
    & 14.1 & 0.349 & 0.63 & 32.7 & 4.49
    & 31.4 & \underline{0.987} & \underline{0.04} & 32.9 & 4.65 \\

    & Retinexformer~\cite{cai2023retinexformer}
    & 1 & 17 & 1.6
    & 16.9 & \textbf{0.567} & 0.54 & 28.1 & 4.64
    & 30.3 & 0.983 & 0.05 & 34.0 & \underline{4.47} \\

    & HVI-CIDNet~\cite{yan2025hvi}
    & 1 & 8 & 2.0
    & 15.2 & 0.495 & 0.56 & 23.5 & 4.42
    & \textbf{33.7} & \textbf{0.989} & \textbf{0.02} & 33.1 & 5.03 \\

    & Multinex~\cite{multinex2026}
    & 1 & 3 & 0.04
    & 15.6 & 0.501 & 0.63 & 21.6 & 7.73
    & 24.8 & 0.958 & 0.09 & 32.0 & 4.72 \\

    \midrule

    \multirow{3}{*}{Generative}
    & PyDiff~\cite{zhou2023pydiff}
    & 4 & 473 & 97.9
    & 17.3 & 0.544 & 0.56 & 33.2 & 7.38
    & 22.5 & 0.956 & 0.17 & 32.4 & 4.83 \\

    & WCDM~\cite{wcdm}
    & 10 & 90 & 22.1
    & 17.1 & \underline{0.553} & 0.59 & 25.5 & 5.94
    & 26.8 & 0.957 & 0.16 & 31.7 & \textbf{4.08} \\

    & ReDDiT~\cite{reddit}
    & 2 & 128 & 17.4
    & 14.1 & 0.488 & 0.59 & 30.9 & 7.45
    & 22.2 & 0.924 & 0.25 & \textbf{35.9} & 5.24 \\

    \midrule

    \multirow{3}{*}{Ours}
    & \textbf{\method{}-XL}
    & 1 & 48 & 124.0
    & \textbf{20.3} & 0.548 & \textbf{0.48} & \underline{36.0} & 2.92
    & \underline{31.7} & 0.969 & 0.12 & \underline{35.2} & 4.59 \\

    & \textbf{\method{}-M}
    & 1 & 36 & 21.8
    & 19.5 & 0.531 & \underline{0.50} & 35.5 & \underline{2.86}
    & 30.6 & 0.961 & 0.14 & 35.1 & 4.63 \\

    & \textbf{\method{}-S}
    & 1 & 34 & 9.1
    & \underline{19.7} & 0.516 & 0.52 & \textbf{37.1} & \textbf{2.70}
    & 30.3 & 0.962 & 0.14 & 34.9 & 4.62 \\

    \bottomrule
  \end{tabular}}
  \caption{\textbf{Results on paired low-light benchmarks.}
  We evaluate on LSD-DLL and UHD-LOL4K benchmarks and report image
  quality and efficiency metrics. Best results are shown in
  \textbf{bold} and second-best results are \underline{underlined}.}
  \label{tab:paired_lsd_uhd}
  \vspace{-10pt}
\end{table*}

\subsection{Efficient Latent Architecture}
\label{subsec:architecture}
\noindent\textbf{Compact Latent Representation.}
Operating in latent space is central to the efficiency of \method{}. Instead of applying the flow model to full-resolution RGB images, the VAE encoder maps the input into a lower-dimensional representation with reduced spatial resolution. The learned transport therefore acts on a compact latent grid, lowering the cost of each function evaluation and making low-NFE inference more practical.

The latent space also provides a useful restoration prior. Since the VAE is pretrained on a broad natural-image distribution, its representation captures reusable image structure while suppressing some pixel-level noise and dataset-specific degradation patterns. Prior work has similarly shown that pretrained latent representations can support robust image restoration and domain generalization~\cite{oldphotorestoration,ldrps,thomas2025latent}. As a result, the flow model learns exposure and appearance corrections in a natural-image latent domain rather than relying solely on dataset-specific pixel statistics.
For the autoencoder, inspired by~\cite{elir}, we use Tiny AutoEncoder for Stable Diffusion (TAESD)~\cite{taesd}, a lightweight distilled alternative to the Stable Diffusion VAE (SD-VAE)~\cite{rombach2022latent}. TAESD preserves an SD-VAE-compatible latent interface while substantially reducing the cost of encoding and decoding. In our implementation, replacing SD-VAE with TAESD reduces encoding/decoding GFLOPs by $15\times$ and parameter count by $33\times$, with minimal loss in end-to-end restoration performance. This is important for low-NFE generation, where the flow model requires only one or a few evaluations and the VAE cost becomes a significant part of the total inference budget. 

\begin{table}[h]
  \centering
  \small
  \setlength{\tabcolsep}{3pt}
  \resizebox{\linewidth}{!}{%
  \begin{tabular}{@{}l@{\hspace{10pt}}lccccc@{}}
    \toprule
    Category &
    Method &
    NFEs$\downarrow$ &
    GFLOPs$\downarrow$ &
    Params (M)$\downarrow$ &
    NIQE$\downarrow$ &
    MUSIQ$\uparrow$ \\
    \midrule

    \multirow{4}{*}{Non-Generative}
    & TFFormer~\cite{lsd_dataset}
    & 1 & 34 & 5.9
    & \underline{4.42} & 24.1 \\

    & Retinexformer~\cite{cai2023retinexformer}
    & 1 & 17 & 1.6
    & 7.35 & 23.4 \\

    & HVI-CIDNet~\cite{yan2025hvi}
    & 1 & 8 & 2.0
    & 6.26 & 26.5 \\

    & Multinex~\cite{multinex2026}
    & 1 & 3 & 0.04
    & 6.96 & 27.2 \\

    \midrule

    \multirow{3}{*}{Generative}
    & PyDiff~\cite{zhou2023pydiff}
    & 4 & 473 & 97.9
    & 6.96 & 26.3 \\

    & WCDM~\cite{wcdm}
    & 10 & 90 & 22.1
    & 6.16 & 23.7 \\

    & ReDDiT~\cite{reddit}
    & 2 & 128 & 17.4
    & 7.94 & 26.4 \\

    \midrule

    \multirow{3}{*}{Ours}
    & \textbf{\method{}-XL}
    & 1 & 48 & 124.0
    & \textbf{4.33} & \textbf{28.4} \\

    & \textbf{\method{}-M}
    & 1 & 36 & 21.8
    & 4.49 & \underline{28.3} \\

    & \textbf{\method{}-S}
    & 1 & 34 & 9.1
    & 4.99 & 27.7 \\

    \bottomrule
  \end{tabular}}
  \vspace{-5pt}
  \caption{\textbf{Results on LSD-U.}
  Best results are shown in \textbf{bold} and second-best results are
  \underline{underlined}.}
  \label{tab:lsd_u}\vspace{-10pt}
\end{table}

\noindent\textbf{HDiT two-time flow model.}
We implement the two-time conditioned flow model $F_\theta(z_t,t,s)$ with an Hourglass Diffusion Transformer (HDiT) backbone. HDiT combines the multi-scale efficiency of U-Net-like architectures with the representation capacity of diffusion transformers \cite{hdit}. Unlike standard DiT-style architectures~\cite{dit}, which apply global attention at every resolution and incur $O(n^2)$ attention cost, HDiT restricts global attention to low-resolution stages and uses fixed-window neighborhood attention at higher resolutions, yielding linear $O(n)$ scaling with the number of latent image tokens. This design is useful for GLOW-FM because latent restoration requires both global exposure/color correction and preservation of spatially aligned local content from the degraded input, while keeping the two-time flow model efficient at inference.

\begin{table}[h]
  \centering
  \small
  \setlength{\tabcolsep}{3pt}
  \resizebox{\linewidth}{!}{%
  \begin{tabular}{@{}lcccccc@{}}
    \toprule
    Method & NFEs$\downarrow$ & GFLOPs$\downarrow$ & Params (M)$\downarrow$ &
    PSNR$\uparrow$ & MUSIQ$\uparrow$ & NIQE$\downarrow$ \\
    \midrule

    PyDiff~\cite{zhou2023pydiff}
    & 4 & 473 & 97.9
    & \underline{21.9} & \textbf{74.2} & 5.40 \\

    WCDM~\cite{wcdm}
    & 10 & 90 & 22.1
    & \textbf{26.3} & 66.9 & 4.52 \\

    ReDDiT~\cite{reddit}
    & 2 & 128 & 17.4
    & 20.8 & \underline{71.7} & 5.38 \\

    \midrule

    \textbf{\method{}-XL}
    & 1 & 48 & 124
    & 21.8 & 70.4 & \underline{3.71} \\

    \textbf{\method{}-M}
    & 1 & 36 & 21.8
    & 21.3 & 71.5 & 3.79 \\

    \textbf{\method{}-S}
    & 1 & 34 & 9.1
    & 21.2 & 71.3 & \textbf{3.70} \\

    \bottomrule
  \end{tabular}}\vspace{-5pt}
  \caption{\textbf{Results on LOL-v1.}
  Best results are shown in \textbf{bold} and second-best results are
  \underline{underlined}.}
  \label{tab:lolv1}\vspace{-15pt}
\end{table}

\section{Experiments}
\label{sec:experiments}
\subsection{Experimental Setup}
\begin{table*}[t]
  \centering
  \scriptsize
  \setlength{\tabcolsep}{3pt}
  \resizebox{\textwidth}{!}{%
  \begin{tabular}{@{}l@{\hspace{10pt}}lcccccccccccccc@{}}
    \toprule
    \multirow{2}{*}{Category} &
    \multirow{2}{*}{Method} &
    \multirow{2}{*}{GFLOPs$\downarrow$} &
    \multirow{2}{*}{Params (M)$\downarrow$} &
    \multicolumn{2}{c}{\textbf{DICM}} &
    \multicolumn{2}{c}{\textbf{LIME}} &
    \multicolumn{2}{c}{\textbf{MEF}} &
    \multicolumn{2}{c}{\textbf{NPE}} &
    \multicolumn{2}{c}{\textbf{VV}} &
    \multicolumn{2}{c}{\textbf{Mean}} \\
    \cmidrule(lr){5-6}
    \cmidrule(lr){7-8}
    \cmidrule(lr){9-10}
    \cmidrule(lr){11-12}
    \cmidrule(lr){13-14}
    \cmidrule(l){15-16}
    & & & &
    NIQE$\downarrow$ & BRISQUE$\downarrow$ &
    NIQE$\downarrow$ & BRISQUE$\downarrow$ &
    NIQE$\downarrow$ & BRISQUE$\downarrow$ &
    NIQE$\downarrow$ & BRISQUE$\downarrow$ &
    NIQE$\downarrow$ & BRISQUE$\downarrow$ &
    NIQE$\downarrow$ & BRISQUE$\downarrow$ \\
    \midrule

    \multirow{9}{*}{Non-Generative}
    & SNR-Aware~\cite{snr_aware}
    & 24 & 39.1
    & 6.17 & 33.2
    & 5.96 & 34.9
    & 4.33 & 27.9
    & 6.51 & 36.9
    & 11.89 & 80.0
    & 6.97 & 42.6 \\

    & DCC-Net~\cite{zhang2022deep}
    & 64 & 13.2
    & 3.44 & 27.0
    & 4.26 & 25.4
    & 4.13 & 30.2
    & 3.51 & 25.1
    & 3.55 & 26.2
    & 3.78 & 26.8 \\

    & Zero-DCE~\cite{guo2020zero}
    & 5 & 0.079
    & 2.84 & 19.3
    & 3.80 & 18.2
    & 4.02 & 26.3
    & 3.59 & 26.4
    & 3.21 & 29.1
    & 3.49 & 23.8 \\

    & KinD++~\cite{zhang2021beyond}
    & 371 & 8.3
    & 3.10 & 20.5
    & 4.72 & 23.3
    & 4.20 & 34.1
    & 3.79 & 27.6
    & 2.68 & 26.5
    & 3.70 & 26.4 \\

    & CPGA-Net~\cite{cpga_net}
    & 2 & 0.025
    & 3.62 & 28.1
    & 4.59 & 21.5
    & 4.18 & 28.1
    & 3.73 & 26.4
    & 4.22 & 35.3
    & 4.07 & 27.9 \\

    & HVI-CIDNet~\cite{yan2025hvi}
    & 8 & 2.0
    & 3.02 & 19.0
    & 3.82 & 14.2
    & 3.82 & 14.1
    & 3.43 & 22.5
    & 3.25 & 23.3
    & 3.47 & 18.6 \\

    & Retinexformer~\cite{cai2023retinexformer}
    & 17 & 1.6
    & 3.37 & 9.5
    & 3.86 & 12.5
    & 3.82 & 14.9
    & 3.35 & 16.4
    & 2.73 & 14.6
    & 3.42 & 13.6 \\

    & TFFormer~\cite{lsd_dataset}
    & 34 & 5.9
    & 4.45 & 26.3
    & 4.72 & 22.0
    & 4.80 & 25.7
    & 4.56 & 25.9
    & 2.91 & \underline{9.1}
    & 4.29 & 21.8 \\

    & Multinex~\cite{multinex2026}
    & 3 & 0.045
    & 3.14 & 12.0
    & 3.84 & 14.2
    & 3.72 & 16.2
    & 3.37 & 17.6
    & 2.93 & 17.8
    & 3.40 & 15.6 \\

    \midrule

    \multirow{6}{*}{Generative}
    & GSAD~\cite{gsad}
    & 1277 & 17.2
    & 2.93 & 18.1
    & 3.61 & 13.3
    & 3.58 & 14.6
    & 3.37 & 21.2
    & 3.22 & 23.9
    & 3.34 & 18.2 \\

    & PyDiff~\cite{zhou2023pydiff}
    & 473 & 97.9
    & 3.86 & 13.4
    & 4.35 & 18.7
    & 5.19 & 28.2
    & 3.68 & 20.8
    & 3.52 & 19.0
    & 4.12 & 20.0 \\

    & ReDDiT~\cite{reddit}
    & 128 & 17.4
    & 3.70 & 15.5
    & 3.80 & 12.8
    & 3.88 & 11.9
    & 3.59 & 14.6
    & 2.65 & 13.9
    & 3.53 & 13.7 \\

    & WCDM~\cite{wcdm}
    & 90 & 22.1
    & 3.36 & 15.8
    & 3.59 & 14.4
    & 3.78 & 16.7
    & 3.37 & 18.3
    & 2.35 & \textbf{8.0}
    & 3.29 & 14.7 \\

    & Reti-Diff~\cite{he2025reti}
    & 175 & 26.1
    & 3.10 & \textbf{7.8}
    & 3.66 & 12.0
    & 3.68 & 15.33
    & 3.39 & 17.0
    & 2.88 & 16.8
    & 3.34 & 13.8 \\

    & LightenDiffusion~\cite{Jiang_2024_ECCV}
    & 92 & 27.8
    & 3.37 & 19.5
    & 4.02 & 17.1
    & 4.00 & 23.8
    & 3.33 & 20.8
    & 3.43 & 27.4
    & 3.63 & 21.7 \\

    \midrule

    \multirow{3}{*}{Ours}
    & \textbf{\method{}-XL}
    & 48 & 124.0
    & 2.61 & \underline{7.9}
    & \textbf{3.12} & \textbf{8.9}
    & 3.28 & 10.9
    & \underline{2.94} & \underline{12.5}
    & \textbf{2.28} & 13.7
    & 2.86 & \textbf{10.8} \\

    & \textbf{\method{}-M}
    & 36 & 21.8
    & \underline{2.60} & 8.2
    & 3.17 & 10.1
    & \textbf{3.19} & \underline{10.8}
    & \textbf{2.93} & 12.6
    & 2.32 & 14.0
    & \underline{2.84} & 11.1 \\

    & \textbf{\method{}-S}
    & 34 & 9.1
    & \textbf{2.58} & \textbf{7.8}
    & \underline{3.16} & \underline{9.7}
    & \underline{3.21} & \textbf{10.7}
    & 2.95 & \textbf{12.1}
    & \underline{2.30} & 14.3
    & \textbf{2.83} & \underline{10.9} \\

    \bottomrule
  \end{tabular}}\vspace{-7pt}
  \caption{\textbf{Cross-dataset generalization benchmark.}
  Results on DICM, LIME, MEF, NPE, and VV. Lower is better.
  Best results are shown in \textbf{bold} and second-best results
  are \underline{underlined}.}
  \label{tab:unpaired_niqe}
  \vspace{-15pt}
\end{table*}

\noindent\textbf{Datasets.}
We evaluate \method{} on both paired and unpaired real-world LLIE benchmarks. Our primary benchmark is LSD-DLL~\cite{lsd_dataset}, a large-scale dataset designed to improve the realism and diversity of LLIE evaluation. Unlike many earlier low-light datasets~\cite{lolv1,lolv2,lsrw}, which are limited by synthetic darkening, controlled capture settings, or restricted sensor coverage, LSD-DLL~\cite{lsd_dataset} provides real low and normal-light image pairs captured across diverse scenes and devices in the wild using dynamic exposure calibration and multi-frame alignment to obtain more realistic paired data. As a result, it better captures phenomena such as sensor noise, spatially varying illumination, and color distortions that are often underrepresented in existing LLIE benchmarks. These properties motivated its use as the benchmark dataset in~\cite{Ciubotariu_2026_CVPRW}. LSD-DLL also provides the LSD-U test split, which consists of unpaired low-light images captured with unseen devices and enables evaluation of generalization beyond the paired training distribution. To complement LSD-DLL, we additionally evaluate on UHD-LOL4K~\cite{uhd_lol_4k_dataset}, a high-resolution LLIE benchmark, and LOL-v1~\cite{lolv1}, one of the most widely used datasets in the LLIE literature.

Following~\cite{reddit,gsad,wcdm}, we further evaluate cross-dataset generalization by training on LOL-v2-Syn~\cite{lolv2} and testing directly on five widely used unpaired real-world low-light datasets: DICM~\cite{dcim_dataset}, LIME~\cite{lime_dataset}, MEF~\cite{mef_dataset}, NPE~\cite{npe_dataset}, and VV~\cite{vv_dataset}. Since these datasets do not provide paired reference images, they are evaluated using no-reference image quality metrics.

\noindent\textbf{Evaluation metrics.}
For paired benchmarks, we evaluate both distortion and perceptual quality.
Distortion is measured using PSNR and SSIM~\cite{ssim_metric}, while perceptual similarity is assessed using LPIPS~\cite{lpips_metric}. We additionally evaluate perceptual image quality using the no-reference metrics NIQE~\cite{niqe_metric} and MUSIQ~\cite{musiq_metric}. For unpaired benchmarks, where ground-truth reference images are unavailable, we follow~\cite{wcdm} and report the no-reference perceptual quality metrics NIQE~\cite{niqe_metric} and BRISQUE~\cite{brisque_metric}. Results are averaged over five evaluation seeds; full mean and STD results are provided in \cref{sec:supp_quantitative}.

\noindent\textbf{Implementation details.}
\method{} employs a frozen Tiny-VAE~\cite{taesd} to obtain 16-channel latent image representations, while only the latent flow model is optimized during training. Unless otherwise specified, experiments use the \method{}-XL configuration with a single NFE. We additionally evaluate medium (M) and small (S) configurations, which differ only in the capacity of the HDiT backbone. Source latents are perturbed with Gaussian noise using $\sigma=0.01$. Models are trained for approximately 100k optimization steps on 8 NVIDIA H100 GPUs.
For LSD-DLL and UHD-LOL4K, we trained the baselines on the respective training sets, whereas all other results use the authors' official code and checkpoints. \cref{sec:supp_baseline_training} details these
baseline training protocols together with our implementation, hyperparameters, and runtime and memory analyses.

\subsection{Results on Paired Benchmarks}
\cref{tab:paired_lsd_uhd} compares \method{} with representative
non-generative and recent efficient generative LLIE methods on the paired
LSD-DLL and UHD-LOL4K benchmarks. Across both datasets, \method{} achieves
a favorable quality--efficiency trade-off with only a single NFE during
inference. On LSD-DLL, it obtains the strongest overall performance across
several distortion and perceptual-quality metrics. On UHD-LOL4K,
non-generative methods remain stronger on some distortion-oriented metrics,
while \method{} maintains competitive image quality with substantially lower
inference cost than the generative baselines. The medium and small variants
further reduce computational cost while preserving strong image quality.
Representative qualitative results on LSD-DLL are shown in \cref{fig:lsd_dll_qualitative}.

On LOL-v1, \method{} maintains a strong quality-efficiency tradeoff. As shown in~\cref{tab:lolv1}, \method{} achieves the best NIQE score among the compared methods with a single NFE. We hypothesize that its weaker distortion performance arises from the limited training data available in this benchmark. A controlled data-scaling experiment supporting this interpretation is presented in \cref{fig:dataset_size_ablation}.
Since paired evaluation does not fully capture robustness to unseen devices and low-light domains, we separately examine cross-dataset and unseen-device generalization in \cref{sec:generalization}.\vspace{-3pt}

\subsection{Generalization to Unseen Low-Light Data} \label{sec:generalization}
\begin{figure*}[t]
  \centering
  \includegraphics[width=0.9\textwidth]{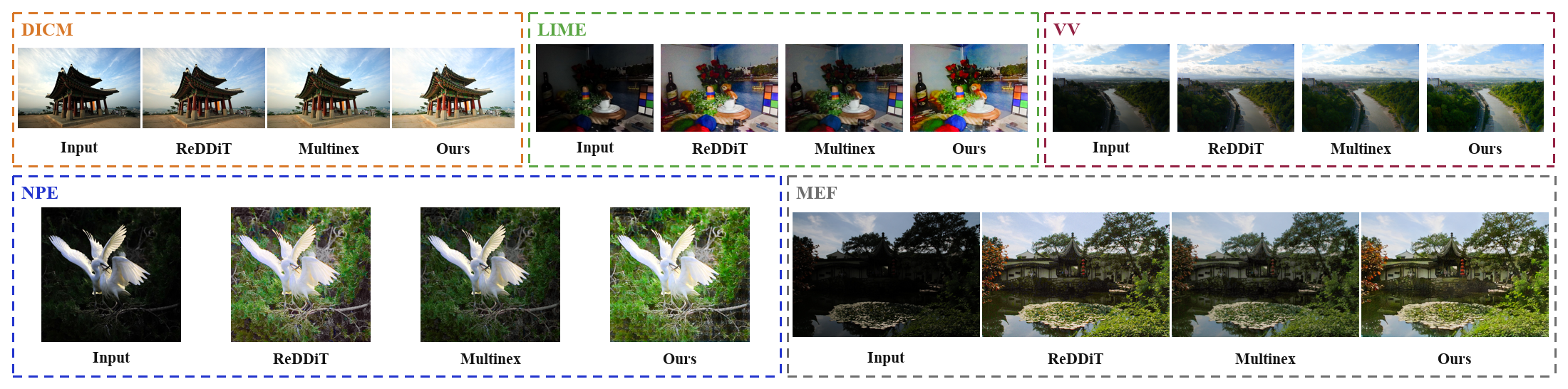}\vspace{-5pt}
  \caption{\textbf{Qualitative generalization comparison.} Results on unpaired real-world low-light benchmarks show that \method{} recovers natural illumination while preserving scene structure across diverse inputs.}\vspace{-20pt}
  \label{fig:qualitative_unpaired}
\end{figure*}

\cref{tab:unpaired_niqe} reports results on five widely used unpaired real-world low-light benchmarks: DICM, LIME, MEF, NPE, and VV. Following the protocol adopted in prior work~\cite{reddit,gsad}, \method{} is trained on LOL-v2-Syn~\cite{lolv2} and evaluated directly on the target datasets without additional fine-tuning.
\method{} achieves the best generative NIQE and BRISQUE results on
nearly all evaluated benchmarks while operating at substantially lower
computational cost. It also obtains the strongest mean performance
across both metrics.

\cref{tab:lsd_u} reports results on the LSD-U subset from~\cite{Ciubotariu_2026_CVPRW}, which evaluates generalization to low-light images captured with unseen devices. Among all compared
generative and non-generative methods, \method{} achieves the best NIQE and MUSIQ scores with a single NFE. Moreover, all \method{} variants achieve competitive no-reference image quality despite substantially lower computational cost than the competing generative methods. These results demonstrate strong generalization beyond the training distribution, indicating that the proposed latent flow formulation remains robust under real-world distribution shifts. Together with the cross-dataset results in \cref{tab:unpaired_niqe}, they demonstrate consistent generalization across unseen devices and diverse low-light domains.

Given the perception--distortion trade-off~\cite{blau2018perception}, we evaluate metrics from both families. Quantitative results and paired ground-truth comparisons in \cref{fig:lsd_dll_qualitative} indicate \method{} offers a favorable balance among the compared methods, while unpaired examples in \cref{fig:qualitative_unpaired} show natural illumination and fewer artifacts; additional examples are provided in \cref{sec:supp_qualitative}.
\vspace{-3pt}
\subsection{Ablation Study}
\label{sec:ablation_study}\vspace{-5pt}
\noindent\textbf{Comparison of one-step objectives.}
\cref{tab:architecture_ablation_validation} compares TVM with
MeanFlow~\cite{meanflow}, ConsistencyFM~\cite{cfm} and No-TVM (standard Flow Matching \cite{lipman2023flow, rectifiedflow}), with all other experimental settings held fixed. We additionally evaluate No-TVM at 25 NFEs. Among the matched TAESD one-step variants, TVM achieves the strongest overall distortion and perceptual performance.
MeanFlow forms its JVP target from the initial-time velocity, whose
sample-wise approximation can destabilize training~\cite{tvm}; this may be particularly problematic for LLIE,
where the initial state is a degraded low-light image. ConsistencyFM enforces a velocity-consistent
straight flow, which may restrict a single full-interval mapping for
spatially heterogeneous restoration~\cite{cfm,elir}. TVM instead
regularizes the displacement at its predicted terminal state, aligning
the one-step mapping with the velocity near the enhanced
endpoint~\cite{tvm}.

\begin{figure}[t]
  \centering
  \includegraphics[width=\linewidth, trim=0.5cm 0cm 0.55cm 0.2cm,
  clip]{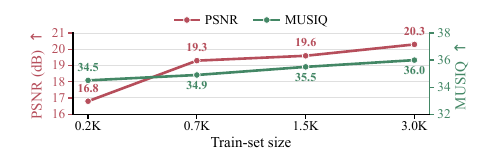}\vspace{-12pt}
    \caption{\textbf{Effect of train-set size.} Results on LSD-DLL for models trained with increasing train-set size.}\vspace{-15pt}
  \label{fig:dataset_size_ablation}
\end{figure}

\noindent\textbf{Effect of training-set size.}
\label{sec:ablation_dataset_size}
\cref{fig:dataset_size_ablation} isolates the effect of data availability
by training \method{} on progressively larger LSD-DLL subsets while
keeping the architecture and training protocol fixed. Increasing the
training set from approximately $0.2\mathrm{K}$ to $3.0\mathrm{K}$
image pairs improves PSNR monotonically from $16.8$ to
$20.3$\,dB and MUSIQ from $34.5$ to $36.0$. This trend supports our
data-limited interpretation of the LOL-v1 distortion results and is
consistent with prior observations that flow-based models benefit
strongly from additional training data~\cite{zhu2025diffusion}.

\noindent\textbf{Effect of latent representation.}
\cref{tab:architecture_ablation_validation} further evaluates the effect of the latent representation on \method{} performance. We compare the proposed TAESD latent representation against a larger SD-VAE and a direct pixel-space variant while keeping the velocity network fixed.
TAESD achieves better PSNR, SSIM, and NIQE with nearly an order of magnitude less computation, whereas SD-VAE achieves better LPIPS and MUSIQ at substantially higher computational and parameter costs. Additional autoencoder-reconstruction analysis is provided in \cref{sec:supp_ablation}. We further compare against a pixel-space variant that removes the latent representation entirely. Despite requiring significantly more computation, direct pixel-space modeling produces substantially weaker distortion and perceptual quality. These results indicate that the proposed latent-space formulation is not only computationally advantageous but also improves enhancement quality, suggesting that the compact TAESD representation provides an effective latent space for LLIE. \vspace{-7pt}

\begin{table}[t]
  \centering
  \setlength{\tabcolsep}{3pt}
  \resizebox{\linewidth}{!}{%
  \begin{tabular}{@{}lcccccccc@{}}
    \toprule
    Variant & Latent & GFLOPs$\downarrow$ & Params (M)$\downarrow$ &
    PSNR$\uparrow$ & SSIM$\uparrow$ & LPIPS$\downarrow$ &
    MUSIQ$\uparrow$ & NIQE$\downarrow$ \\
    \midrule

    Pixel Space (TVM) & None & 1693 & 121 &
    16.0 & 0.47 & 0.54 & 28.2 & 4.51 \\

    SD-VAE (TVM) & SD-VAE & 471 & 205 &
    19.6 & 0.50 & 0.47 & \textbf{40.6} & 3.37 \\

    No TVM (1 NFE) & TAESD & 48 & 124 &
    19.7 & 0.50 & 0.48 & 27.7 & 6.00 \\

    No TVM (25 NFE) & TAESD & 415 & 124 &
    20.1 & 0.54 & \textbf{0.37} & 35.6 & 3.00 \\

    ConsistencyFM & TAESD & 48 & 124 &
    19.4 & 0.52 & 0.50 & 32.8 & 3.53 \\

    MeanFlow & TAESD & 48 & 124 &
    19.5 & 0.52 & 0.49 & 31.7 & 3.59 \\

    \method{} & TAESD & 48 & 124 &
    \textbf{20.3} & \textbf{0.55} & 0.48 & 36.0 & \textbf{2.92} \\
    \bottomrule
  \end{tabular}}
  \caption{\textbf{Ablation study.} Results on LSD-DLL assessing the
  latent representation and one-step training objective.}
\label{tab:architecture_ablation_validation}
\end{table}

\section{Conclusion}
\label{sec:conclusion}
Low-light image enhancement remains challenging because noise, color shifts, and information loss make faithful restoration ambiguous. 
Although recent generative LLIE methods have reduced sampling cost, combining single-step inference, low overall computation, and robust performance across real-world conditions remains challenging.

In this work, we address this challenge with \method{}, an efficient generative LLIE framework based on one-step latent flow matching. By combining efficiency-aware latent flow matching with Terminal Velocity Matching, \method{} enables single-NFE enhancement while preserving generative priors for realistic restoration. Experiments show competitive quality, improved efficiency, and state-of-the-art cross-dataset generalization, suggesting a practical direction for robust and efficient generative low-light enhancement.

{
    \small
    \bibliographystyle{ieeenat_fullname}
    \bibliography{main}
}
\newpage
\appendix
\maketitlesupplementary
\section{Supplementary Overview}
This supplementary material provides additional details and results supporting the main paper and is organized as follows:

\begin{itemize}
    \item \cref{app:tvm_w2_bound} provides the pixel-space Wasserstein bound for latent TVM.
    \item \cref{sec:app_implementation_details} presents additional implementation details.
    \item \cref{sec:supp_baseline_training} describes the training protocols used for the baseline methods.
    \item \cref{sec:supp_ablation} provides further ablation studies.
    \item \cref{sec:supp_quantitative} reports additional quantitative results.
    \item \cref{sec:supp_qualitative} presents additional qualitative comparisons.
\end{itemize}
\section{Pixel-Space Wasserstein Bound for Latent TVM}
\label[appendix]{app:tvm_w2_bound}

\paragraph{Motivation, notation, and definitions.}
In the main paper, we motivate TVM not only by its strong one-step generation performance, but also through theoretical analysis. We make this point precise below by showing that the Wasserstein-2 distance~\cite{panaretos2019statistical} between the predicted normal-light image distribution and the true normal-light image distribution is controlled by the latent TVM training objective and the VAE reconstruction error. Thus, minimizing our latent TVM objective reduces an upper bound on the pixel-space distributional gap.

Let $\mathcal{X}$ and $\mathcal{Z}$ denote the image and latent spaces, respectively. Let $x_0\sim p_{x_0}$ be a normal-light target image, and let $z_0=\mathcal{E}(x_0)$ be its latent representation under the frozen VAE encoder $\mathcal{E}$. The decoder is denoted by $\mathcal{D}:\mathcal{Z}\to\mathcal{X}$. The TVM flow model has trainable parameters $\theta$ and produces an estimated normal-light latent $\hat{z}_0\sim p_{\hat{z}_0}^{\theta}$, whose decoded prediction is $\hat{x}_0=\mathcal{D}(\hat{z}_0)$. We also define the VAE reconstruction of the target image as $\tilde{x}_0=\mathcal{D}(\mathcal{E}(x_0))=\mathcal{D}(z_0)$.

We use $\#$ to denote the pushforward of a distribution. Hence,
\begin{equation}
\label{eq:pushforward_eq}
p_{z_0}=\mathcal{E}_{\#}p_{x_0},
\qquad
p_{\hat{x}_0}^{\theta}=\mathcal{D}_{\#}p_{\hat{z}_0}^{\theta},
\qquad
p_{\tilde{x}_0}=\mathcal{D}_{\#}p_{z_0}.
\end{equation}
The variable $z_s$ denotes a latent point at time $s$, $v_s$ denotes the corresponding sample-wise conditional velocity, and $u(z_s,s)$ denotes the marginal latent velocity field. As in Flow Matching~\cite{lipman2023flow}, $u(z_s,s)=\mathbb{E}[v_s\mid z_s]$.

We denote the terminal-velocity matching loss by $\mathcal{L}_{\mathrm{TVM}}^{t,s}(\theta)$ and the conditional flow-matching loss used in our training objective by $\mathcal{L}_{\mathrm{FM}}^s(\theta)$. For compactness, define the combined per-time objective
\begin{equation}
\label{eq:combined_tvm_fm_loss}
\mathcal{L}^{t,s}(\theta)
=
\mathcal{L}_{\mathrm{TVM}}^{t,s}(\theta)
+
\mathcal{L}_{\mathrm{FM}}^s(\theta).
\end{equation}
The VAE reconstruction error is
\begin{equation}
\label{eq:vae_reconstruction_error}
\epsilon_{\mathrm{VAE}}
=
\mathbb{E}_{x_0\sim p_{x_0}}
\|\mathcal{D}(\mathcal{E}(x_0))-x_0\|_2^2.
\end{equation}
For distributions $\mu,\nu$ over a space $\mathcal{Y}$, the Wasserstein-2 distance is
\begin{equation}
\label{eq:w2_definition}
W_{2,\mathcal{Y}}(\mu,\nu)
=
\sqrt{
\inf_{\gamma\in\Pi(\mu,\nu)}
\int_{\mathcal{Y}\times\mathcal{Y}}
\|y-y'\|_2^2\,d\gamma(y,y')
},
\end{equation}
where $\Pi(\mu,\nu)$ is the set of couplings with marginals $\mu$ and $\nu$.

We will prove the following result:

\begin{theorem}[Pixel-space Wasserstein bound for latent TVM]
\label{thm:pixel_w2_tvm_bound}
Assume that the TVM regularity conditions of~\cite[Theorem~1]{tvm} hold in latent space, and the frozen decoder $\mathcal{D}$ is $L_{\mathcal{D}}$-Lipschitz on the relevant latent support. Then
\begin{equation}
W_{2,\mathcal{X}}(p_{\hat{x}_0}^{\theta},p_{x_0})
\le
L_{\mathcal{D}}
\sqrt{
\int_0^t
\lambda[L](s)\mathcal{L}^{t,s}(\theta)
\,ds
}
+
\sqrt{\epsilon_{\mathrm{VAE}}},
\end{equation}
where $\lambda[\cdot]$ is a functional of the Lipschitz constants $L(s)$.
\end{theorem}

Consequently, in the ideal regime where the weighted TVM/FM training objective converges to zero and the VAE reconstruction error is zero, the Wasserstein-2 bound also vanishes. For a fixed pretrained VAE, $\epsilon_{\mathrm{VAE}}$ remains a fixed reconstruction residual, while TVM training minimizes the trainable part of the bound.
    \begin{lemma}[Pixel-Space $W_2$ Bound under a Lipschitz Decoder]
    \label{lem:decoder_pushforward_bound}
     Let $\mathcal{D}$ be $L_{\mathcal{D}}$-Lipschitz on the relevant latent support:
    \begin{equation}\label{eq:lipschitz_eq}
    \|\mathcal{D}(z)-\mathcal{D}(z')\|_2
    \le
    L_{\mathcal{D}}\|z-z'\|_2.
    \end{equation}
    Then
    \begin{equation}
    W_{2,\mathcal{X}}(p_{\hat{x}_0}^{\theta},p_{\tilde{x}_0})
    \le
    L_{\mathcal{D}}
    W_{2,\mathcal{Z}}(p_{\hat{z}_0}^{\theta},p_{z_0}).
    \end{equation}
    \end{lemma}
    
    \begin{proof}
    Following the decoder pushforward argument used in~\cite[Eq.~(11)]{elir},
    we first write the Wasserstein-2 distance in pixel space:
    \begin{equation}
    \begin{aligned}
    &W_{2,\mathcal{X}}(p_{\hat{x}_0}^{\theta},p_{\tilde{x}_0}) \\
    &=
    \sqrt{
    \inf_{\gamma\in\Pi(p_{\hat{x}_0}^{\theta},p_{\tilde{x}_0})}
    \int_{\mathcal{X}\times\mathcal{X}}
    \|\hat{x}_0-\tilde{x}_0\|_2^2
    \,d\gamma(\hat{x}_0,\tilde{x}_0)
    }.
    \end{aligned}
    \end{equation}
    Using the pushforward relations induced by the decoder $\mathcal{D}$ in \cref{eq:pushforward_eq}, together with the expectation form of the Wasserstein-2 distance~\cite{panaretos2019statistical} and the law of the unconscious statistician (LOTUS), the same distance can be expressed over latent-space couplings as
    \begin{equation}
    \begin{aligned}
    &W_{2,\mathcal{X}}(p_{\hat{x}_0}^{\theta},p_{\tilde{x}_0}) \\
    &=
    \sqrt{
    \inf_{\pi\in\Pi(p_{\hat{z}_0}^{\theta},p_{z_0})}
    \int_{\mathcal{Z}\times\mathcal{Z}}
    \|\mathcal{D}(\hat{z}_0)-\mathcal{D}(z_0)\|_2^2
    \,d\pi(\hat{z}_0,z_0)
    }.
    \end{aligned}
    \end{equation}
    
    Using the Lipschitz property of $\mathcal{D}$ from \cref{eq:lipschitz_eq} we get:
    
    \begin{equation}
    \begin{aligned}
    &W_{2,\mathcal{X}}(p_{\hat{x}_0}^{\theta},p_{\tilde{x}_0}) \\
    &\le
    \sqrt{
    L_{\mathcal{D}}^2
    \inf_{\pi\in\Pi(p_{\hat{z}_0}^{\theta},p_{z_0})}
    \int_{\mathcal{Z}\times\mathcal{Z}}
    \|\hat{z}_0-z_0\|_2^2
    \,d\pi(\hat{z}_0,z_0)
    } \\
    &=
    L_{\mathcal{D}}
    W_{2,\mathcal{Z}}(p_{\hat{z}_0}^{\theta},p_{z_0}).
    \end{aligned}
    \end{equation}
    \end{proof}

\begin{lemma}[Constant-Free TVM Bound]
\label{lem:tvm_latent_bound}
Let $p_{\hat{z}_0}^{\theta}$ denote the latent distribution obtained by the TVM pushforward from time $t$ to $0$, and let $p_{z_0}$ denote the target normal-light latent distribution. Assume that the TVM regularity conditions of~\cite[Theorem~1]{tvm} hold in latent space, then
\begin{equation}
W_{2,\mathcal{Z}}^2(p_{\hat{z}_0}^{\theta},p_{z_0})
\le
\int_0^t
\lambda[L](s)
\left[
\mathcal{L}_{\mathrm{TVM}}^{t,s}(\theta)
+
\mathcal{L}_{\mathrm{FM}}^s(\theta)
\right]
\,ds,
\end{equation}
where $\lambda[\cdot]$ is a functional of the Lipschitz constants $L(s)$.
\end{lemma}

\begin{proof}
The TVM proof~\cite[Eq.~(56)]{tvm} first obtains the following latent-space bound:
\begin{equation}
\begin{aligned}
&W_{2,\mathcal{Z}}^2(p_{\hat{z}_0}^{\theta},p_{z_0}) \\
&\le
\int_0^t
\lambda[L](s)
\left[
\mathcal{L}_{\mathrm{TVM}}^{t,s}(\theta)
+
\underbrace{
\mathbb{E}_{z_s}
\left\|
u_{\theta}(z_s,s)-u(z_s,s)
\right\|_2^2
}_{(a)}
\right]ds .
\end{aligned}
\label{eq:tvm_eq56_rewrite}
\end{equation}
The first term is the terminal-velocity matching loss and the second term, denoted by $(a)$, is the marginal flow-matching loss:
\begin{equation}
(a)
=
\mathcal{L}_{\mathrm{MFM}}^s(\theta)
=
\mathbb{E}_{z_s}
\|u_{\theta}(z_s,s)-u(z_s,s)\|_2^2,
\end{equation}
where $u(z_s,s)$ is the marginal velocity field.

To connect $(a)$ to the practical flow-matching objective used in our training, we use the Marginal Flow-Matching (MFM) / Conditional Flow-Matching (CFM) equivalence of Lipman et al.~\cite[Theorem~2]{lipman2023flow}.\footnote{Lipman et al.~\cite{lipman2023flow} denote the marginal objective by FM and the conditional objective by CFM. Here, we denote the marginal objective by MFM and the conditional training objective by FM to remain consistent with the notation used in the main paper.}
In our notation, the conditional flow-matching loss is
\begin{equation}
\mathcal{L}_{\mathrm{FM}}^s(\theta)
=
\mathbb{E}_{z_s,v_s}
\|u_{\theta}(z_s,s)-v_s\|_2^2,
\end{equation}
where the marginal velocity is the conditional expectation of the sample-wise velocity:
\begin{equation}
u(z_s,s)=\mathbb{E}[v_s\mid z_s].
\end{equation}

Following the standard MFM/CFM relation~\cite[Theorem~2]{lipman2023flow}, add and subtract $u(z_s,s)$:
\begin{equation}
u_{\theta}(z_s,s)-v_s
=
u_{\theta}(z_s,s)-u(z_s,s)
+
u(z_s,s)-v_s.
\end{equation}
Expanding the squared norm and taking expectation gives
\begin{align}
\mathcal{L}_{\mathrm{FM}}^s(\theta)
&=
\mathcal{L}_{\mathrm{MFM}}^s(\theta)
+
\mathbb{E}_{z_s,v_s}
\|u(z_s,s)-v_s\|_2^2 \nonumber\\
&\quad+
2\mathbb{E}_{z_s,v_s}
\left[
\left\langle
u_{\theta}(z_s,s)-u(z_s,s),
u(z_s,s)-v_s
\right\rangle
\right].
\end{align}
The cross term vanishes because
\begin{equation}
\mathbb{E}[u(z_s,s)-v_s\mid z_s]
=
u(z_s,s)-\mathbb{E}[v_s\mid z_s]
=
0.
\end{equation}
Therefore,
\begin{equation}
\mathcal{L}_{\mathrm{FM}}^s(\theta)
=
\mathcal{L}_{\mathrm{MFM}}^s(\theta)
+
\mathbb{E}_{z_s,v_s}
\|u(z_s,s)-v_s\|_2^2.
\end{equation}
Since the second term is nonnegative, we obtain
\begin{equation}
(a)
=
\mathcal{L}_{\mathrm{MFM}}^s(\theta)
\le
\mathcal{L}_{\mathrm{FM}}^s(\theta).
\end{equation}

Substituting this inequality into \cref{eq:tvm_eq56_rewrite}, the marginal flow-matching term $(a)$ is upper-bounded by the flow-matching training objective. Hence,
\begin{equation}
W_{2,\mathcal{Z}}^2(p_{\hat{z}_0}^{\theta},p_{z_0})
\le
\int_0^t
\lambda[L](s)
\left[
\mathcal{L}_{\mathrm{TVM}}^{t,s}(\theta)
+
\mathcal{L}_{\mathrm{FM}}^s(\theta)
\right]
\,ds. \nonumber
\end{equation}
\end{proof}

\begin{lemma}[VAE Reconstruction Wasserstein Bound]
\label{lem:vae_reconstruction_bound}
Let $\tilde{x}_0=\mathcal{D}(\mathcal{E}(x_0))$ and let
$p_{\tilde{x}_0}=(\mathcal{D}\circ\mathcal{E})_{\#}p_{x_0}$. Assume the VAE reconstruction error is finite:
\begin{equation}
\epsilon_{\mathrm{VAE}}
=
\mathbb{E}_{x_0\sim p_{x_0}}
\|\mathcal{D}(\mathcal{E}(x_0))-x_0\|_2^2.
\end{equation}
Then
\begin{equation}
W_{2,\mathcal{X}}(p_{\tilde{x}_0},p_{x_0})
\le
\sqrt{\epsilon_{\mathrm{VAE}}}.
\end{equation}
\end{lemma}

\begin{proof}
Following the reconstruction-to-Wasserstein argument used in~\cite[Eq.~(10)]{elir}, we start from the definition of the Wasserstein-2 distance:
\begin{align}
&W_{2,\mathcal{X}}(p_{\tilde{x}_0},p_{x_0})
= \nonumber \\
&\sqrt{
\inf_{\gamma\in\Pi(p_{\tilde{x}_0},p_{x_0})}
\int_{\mathcal{X}\times\mathcal{X}}
\|\tilde{x}_0-x_0\|_2^2
\,d\gamma(\tilde{x}_0,x_0)
}.
\end{align}
Consider the particular coupling induced by sampling $x_0\sim p_{x_0}$ and setting
\begin{equation}
\tilde{x}_0=\mathcal{D}(\mathcal{E}(x_0)).
\end{equation}
Equivalently, this coupling is
\begin{equation}
\gamma_{\mathrm{VAE}}
=
(\mathcal{D}\circ\mathcal{E},\mathrm{id})_{\#}p_{x_0},
\end{equation}
and it belongs to $\Pi(p_{\tilde{x}_0},p_{x_0})$. Since the infimum is no larger than the cost of any particular coupling,
\begin{align}
&W_{2,\mathcal{X}}(p_{\tilde{x}_0},p_{x_0})
\le \nonumber \\
&\sqrt{
\int_{\mathcal{X}\times\mathcal{X}}
\|\tilde{x}_0-x_0\|_2^2
\,d\gamma_{\mathrm{VAE}}(\tilde{x}_0,x_0)
}.
\end{align}
Using the definition of $\gamma_{\mathrm{VAE}}$, we can substitute
$\tilde{x}_0=\mathcal{D}(\mathcal{E}(x_0))$ and marginalize over $\tilde{x}_0$:
\begin{align}
&\int_{\mathcal{X}\times\mathcal{X}}
\|\tilde{x}_0-x_0\|_2^2
\,d\gamma_{\mathrm{VAE}}(\tilde{x}_0,x_0) \nonumber \\
&=
\int_{\mathcal{X}}
\|\mathcal{D}(\mathcal{E}(x_0))-x_0\|_2^2
\,dp_{x_0}(x_0) \nonumber \\
&=
\mathbb{E}_{x_0\sim p_{x_0}}
\left[
\|\mathcal{D}(\mathcal{E}(x_0))-x_0\|_2^2
\right].
\end{align}
Therefore,
\begin{equation}
\begin{aligned}
W_{2,\mathcal{X}}(p_{\tilde{x}_0},p_{x_0})
&\le
\sqrt{
\mathbb{E}_{x_0\sim p_{x_0}}
\|\mathcal{D}(\mathcal{E}(x_0))-x_0\|_2^2
} \\
&=
\sqrt{\epsilon_{\mathrm{VAE}}}.
\end{aligned}
\end{equation}
\end{proof}

\begin{proof}[\textbf{Proof of \cref{thm:pixel_w2_tvm_bound}}]
By the triangle inequality for the Wasserstein-2 metric,
\begin{equation}
W_{2,\mathcal{X}}(p_{\hat{x}_0}^{\theta},p_{x_0})
\le
W_{2,\mathcal{X}}(p_{\hat{x}_0}^{\theta},p_{\tilde{x}_0})
+
W_{2,\mathcal{X}}(p_{\tilde{x}_0},p_{x_0}).
\end{equation}
Using \cref{lem:decoder_pushforward_bound},
\begin{equation}
W_{2,\mathcal{X}}(p_{\hat{x}_0}^{\theta},p_{\tilde{x}_0})
\le
L_{\mathcal{D}}
W_{2,\mathcal{Z}}(p_{\hat{z}_0}^{\theta},p_{z_0}).
\end{equation}
Using \cref{lem:tvm_latent_bound},
\begin{equation}
W_{2,\mathcal{Z}}(p_{\hat{z}_0}^{\theta},p_{z_0})
\le
\sqrt{
\int_0^t
\lambda[L](s)\mathcal{L}^{t,s}(\theta)
\,ds
}.
\end{equation}
Thus,
\begin{equation}
W_{2,\mathcal{X}}(p_{\hat{x}_0}^{\theta},p_{\tilde{x}_0})
\le
L_{\mathcal{D}}
\sqrt{
\int_0^t
\lambda[L](s)\mathcal{L}^{t,s}(\theta)
\,ds
}.
\end{equation}
Using \cref{lem:vae_reconstruction_bound},
\begin{equation}
W_{2,\mathcal{X}}(p_{\tilde{x}_0},p_{x_0})
\le
\sqrt{\epsilon_{\mathrm{VAE}}}.
\end{equation}
Substituting both bounds into the triangle inequality gives
\begin{equation}
\begin{aligned}
& W_{2,\mathcal{X}}(p_{\hat{x}_0}^{\theta},p_{x_0}) \\
&\quad \le
L_{\mathcal{D}}
\sqrt{
\int_0^t
\lambda[L](s)\mathcal{L}^{t,s}(\theta)
\,ds
}
+
\sqrt{\epsilon_{\mathrm{VAE}}}.
\end{aligned}
\end{equation}
This proves the theorem.
\end{proof}
\section{Additional Implementation Details}
\label[appendix]{sec:app_implementation_details}
\paragraph{Training and inference.}
Unless otherwise specified, all experiments use a frozen Tiny-VAE encoder/decoder while optimizing only the latent flow model. Models are trained on 8 NVIDIA H100 GPUs with a per-GPU batch size of 2 (effective batch size 16) for approximately 100k optimization steps. Optimization uses AdamW with a learning rate of $10^{-5}$, betas $(0.9,0.95)$, weight decay $0.01$, gradient clipping at norm $1.0$, and EMA decay $0.9999$. The learning rate follows a 5k-step warmup with cosine decay to a minimum learning-rate ratio of $0.01$. Training uses random $2048\times2048$ crops for LSD-DLL and UHD-LOL4K, and $384\times384$ crops for LOL-v1. Validation uses a batch size of 1 and is performed at the original image resolution. Source latents are perturbed with Gaussian noise using $\sigma=0.01$ during training and $\sigma=0$ during evaluation. Unless otherwise specified, all reported results use a single NFE.

\paragraph{HDiT configurations.}
\cref{tab:hdit_configurations} summarizes the implementation details of the HDiT backbone variants used in our experiments. Each HDiT backbone has two resolution levels: the first level uses $7{\times}7$ neighborhood attention (NA), and the second level uses global self-attention. In the table, \emph{Widths} and \emph{Depths} denote the channel width and the number of transformer blocks specified at each HDiT level, respectively; \emph{Map. width} and \emph{Map. depth} denote the width and depth of the conditioning/mapping network, and $d_{\mathrm{head}}$ denotes the attention head dimension.

\begin{table*}[t]
\centering
\small
\setlength{\tabcolsep}{5pt}
\resizebox{\textwidth}{!}{%
\begin{tabular}{lcccccccc}
\toprule
Model & Input channels & Widths & Depths & Map. width & Map. depth
& Patch & $d_{\mathrm{head}}$ & NA kernel \\
\midrule
\method{}-XL & 16 & $(384,768)$ & $(2,11)$ & 768 & 2
& $2{\times}2$ & 64 & $7{\times}7$ \\
\method{}-M  & 16 & $(192,384)$ & $(2,6)$  & 384 & 2
& $2{\times}2$ & 64 & $7{\times}7$ \\
\method{}-S  & 16 & $(128,256)$ & $(2,4)$  & 256 & 2
& $2{\times}2$ & 64 & $7{\times}7$ \\
\bottomrule
\end{tabular}}
\caption{HDiT configurations used by \method{}.}
\label{tab:hdit_configurations}
\end{table*}

\paragraph{Tiny-VAE latent representation.}
We use the official \texttt{madebyollin/taesd3}~\cite{taesd} checkpoint as a frozen Tiny-VAE image tokenizer. TAESD is a lightweight autoencoder distilled from the Stable Diffusion VAE (SD-VAE)~\cite{rombach2022latent}, providing a compact alternative for encoding and decoding images. RGB images are encoded into 16-channel latent representations, and the flow model operates entirely in this latent space. The encoder and decoder remain frozen throughout training, and no additional latent normalization is applied. During inference, the predicted latent representation is decoded using the same frozen Tiny-VAE decoder before computing image-space evaluation metrics.

\paragraph{TVM warmup schedule.}
The TVM loss weight is introduced gradually during training according to
\begin{equation}
\mathcal{L}(k)
=
\mathcal{L}_{\mathrm{FM}}
+
\lambda(k)\mathcal{L}_{\mathrm{TVM}},
\end{equation}
where the step-dependent TVM weight is defined as
\begin{equation}
\lambda(k)=
\begin{cases}
0, & k < W, \\
\lambda_{\max}\dfrac{k-W}{R}, & W \leq k < W+R, \\
\lambda_{\max}, & k \geq W+R.
\end{cases}
\end{equation}
Here, $k$ denotes the optimizer step, $W$ the warmup duration, $R$ the linear ramp length, and $\lambda_{\max}$ the final TVM weight. The TVM objective is disabled during warmup, introduced linearly over the ramp period, and kept fixed at $\lambda_{\max}$ thereafter. We use $(W,R,\lambda_{\max})=(5\mathrm{k},10\mathrm{k},1.0)$ for LSD-DLL and $(10\mathrm{k},20\mathrm{k},1.0)$ for UHD-LOL4K.

\section{Baseline Methods: Training Details}
\label[appendix]{sec:supp_baseline_training}

For LSD-DLL and UHD-LOL4K, we independently retrained every baseline on the corresponding training set using the authors' official code and method-specific training recipe. We retained the official architecture, optimization settings, and training schedule, modifying only the dataset configuration as necessary. For LOL-v1, we used the official pretrained weights released by the respective authors whenever available. TFFormer was the only exception: because no official LOL-v1 checkpoint was available, we also trained it from scratch on the LOL-v1 training set. Unless otherwise stated, the training details below describe our retraining on LSD-DLL and UHD-LOL4K, and all models were trained using NVIDIA H100 GPUs.

\paragraph{TFFormer.}
We trained a separate TFFormer model from scratch on each of LSD-DLL, UHD-LOL4K, and LOL-v1. TFFormer was optimized using an Adam optimizer with hyperparameters \(\beta_1=0.9\) and \(\beta_2=0.99\). We set the initial learning rate to \(10^{-4}\) and adjusted it after 100{,}000 steps using PyTorch's ReduceLROnPlateau scheduler~\cite{pytorch_reduce_lr_on_plateau}. We trained each model for 300{,}000 steps on a single GPU using a batch size of 24 and \(128\times128\) RGB patches. Additionally, we used a low-light regularization coefficient of \(\lambda_R=0.2\) for the combined reconstruction and luminance--chrominance objective.

\paragraph{HVI-CIDNet.}
HVI-CIDNet was trained with the Adam optimizer $(\beta_1 = 0.9,\ \beta_2 = 0.999)$ and  \(256\times256\) input crops using a single GPU for 1{,}000 epochs with a batch size of 4. The initial learning rate was \(10^{-4}\) and is then steadily decreased to $1 \times 10^{-7}$ using a cosine annealing scheme~\cite{loshchilov2016sgdr} during training.
We used the default HVI-CIDNet data augmentation pipeline, consisting of random cropping, random horizontal flipping, and random vertical flipping.

\paragraph{Retinexformer.}
We trained Retinexformer using the Adam optimizer with $\beta_1=0.9$ and $\beta_2=0.999$, an initial learning rate of $2\times10^{-4}$ then steadily
decreased to $1\times10^{-6}$ by the cosine annealing scheme~\cite{loshchilov2016sgdr}. Training is set for $2.5\times10^{5}$ iterations. We train with batch size 8 on randomly cropped $128\times128$ patches, and apply data augmentation of random rotation and flipping.

\paragraph{Multinex.}
We trained Multinex using the Adam optimizer with a cosine annealing learning-rate schedule~\cite{loshchilov2016sgdr} that decays from $2\times10^{-4}$ to $1\times10^{-6}$.
Multinex is trained from scratch for 150K iterations with a batch size of 8 and patch size of \(256\times256\). Training data are augmented via random cropping, horizontal and vertical flipping, and random rotation.

\paragraph{PyDiff.}
PyDiff was trained from scratch using random $192\times288$ crops, horizontal flipping, and brightness augmentation with a factor sampled from $[0.5,1.5]$. We used Adam with an initial learning rate of $10^{-4}$ and a batch size of 8 per GPU. The learning rate was multiplied by $0.5$ after 50k, 75k, 100k, 150k, and 200k iterations. Training was conducted for 500k iterations using the official 2,000-step linear diffusion schedule, with variance endpoints of $10^{-6}$ and $10^{-2}$.

\paragraph{WCDM.}
WCDM was trained from scratch using random $256\times256$ crops and a batch size of 12. The model was optimized with Adam using a learning rate of $10^{-4}$ and zero weight decay. The learning rate was multiplied by $0.8$ every 5,000 iterations, and an exponential moving average with a decay of $0.999$ was maintained during training. Training was conducted for 1,000 epochs, following the official training schedule. We used the official 200-step linear diffusion schedule with variance endpoints of $10^{-4}$ and $2\times10^{-2}$.

\paragraph{ReDDiT.}
For each of LSD-DLL and UHD-LOL4K, we followed the ReDDiT progressive-distillation procedure. We first trained the required dataset-specific GSAD teacher from scratch for 750 epochs in FP32 using random \(96\times96\) crops. Training was performed across eight GPUs with a batch size of 8 per GPU, yielding an effective batch size of 64. We used Adam with a learning rate of \(10^{-4}\). The teacher objective combined the \(\ell_1\) reconstruction loss and the GSAD structural loss with weights of 10 and 1, respectively, and employed a 500-step linear diffusion schedule.

Starting from the dataset-specific teacher, we first trained an eight-step ReDDiT model and then progressively distilled it into four-step and two-step students. Each distillation stage was trained for \(5{,}000\) iterations using random \(96\times96\) crops, a batch size of 16, Adam with a learning rate of \(10^{-4}\), and an exponential moving average with a decay of \(0.9999\). The ReDDiT stages used a fine-grained diffusion-time discretization of \(T=512\) with a linear \(\beta\)-schedule whose endpoints were \(10^{-4}\) and \(2\times10^{-2}\).
\section{Additional Ablation Studies}
\label[appendix]{sec:supp_ablation}

\begin{figure}[t]
    \centering
    \includegraphics[width=\linewidth]
        {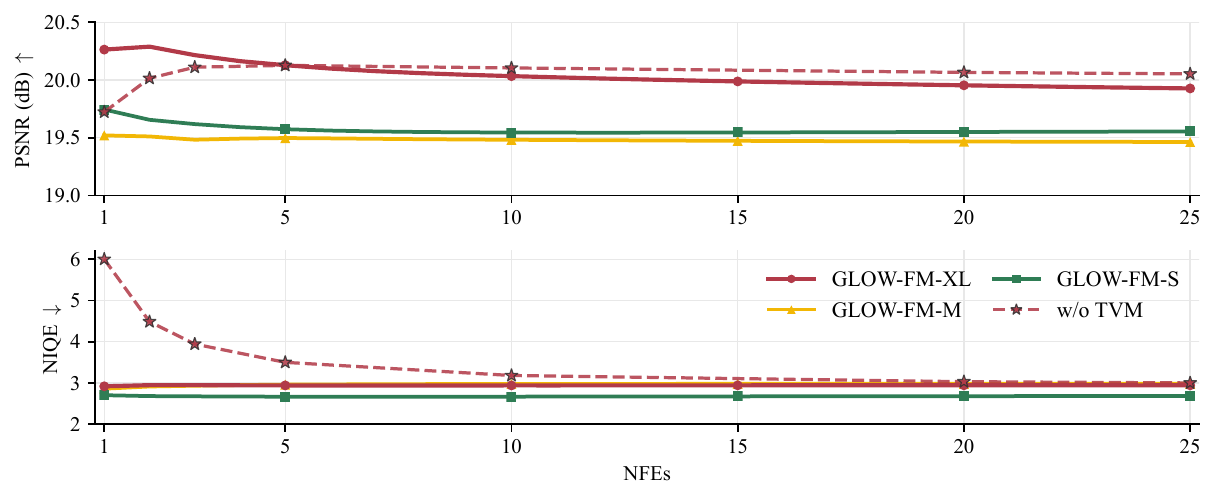}
    \caption{\textbf{NFE ablation.}
    PSNR and NIQE on LSD-DLL as the number of function evaluations
    varies for \method{} with and without TVM.}
    \label{fig:nfe_ablation}
\end{figure}

\begin{figure}[t]
    \centering
    \includegraphics[width=\linewidth]
        {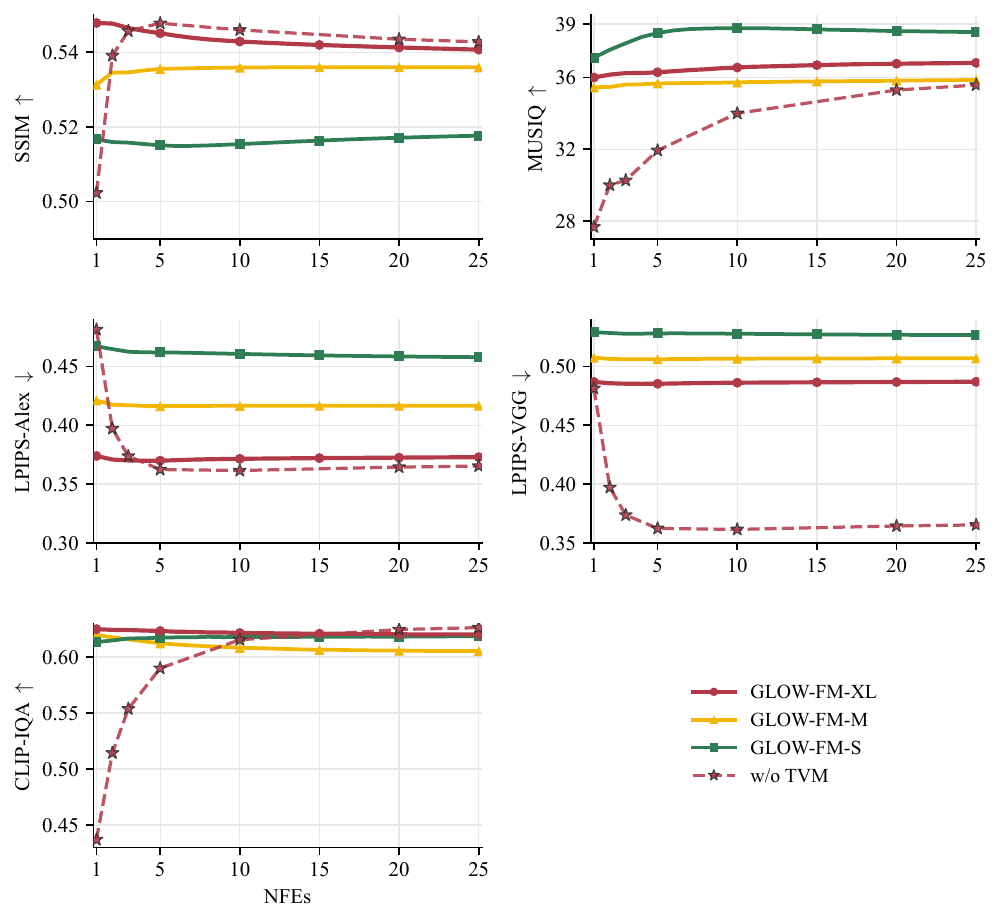}
    \caption{\textbf{Extended NFE ablation.}
    SSIM, LPIPS-VGG, LPIPS-Alex, MUSIQ, and CLIP-IQA on the LSD-DLL
    validation split as the number of function evaluations varies.}
    \label{fig:nfe_add_ablation}
\end{figure}

\paragraph{Extended NFE Ablation.}
\Cref{fig:nfe_ablation,fig:nfe_add_ablation} analyze the effect of the
number of function evaluations (NFEs). \Cref{fig:nfe_ablation} reports
PSNR and NIQE, while \cref{fig:nfe_add_ablation} extends the evaluation
to SSIM, LPIPS-VGG, LPIPS-Alex, MUSIQ, and CLIP-IQA. Across these
metrics, the TVM-trained variants remain largely stable as the number
of NFEs increases, whereas No-TVM improves substantially from one to
25 NFEs. The largest improvement is observed for LPIPS-VGG, for which
No-TVM surpasses the TVM-trained model after only a few NFEs. The
remaining metrics, including LPIPS-Alex, converge toward performance
comparable to that of the TVM-trained model operating with a single
NFE. As reported in \cref{tab:architecture_ablation_lsdu_supp},
No-TVM at 25 NFEs achieves lower LPIPS; however, one-step TVM retains
slightly better PSNR and NIQE with an $8.6\times$ reduction in GFLOPs.
Overall, these results are consistent with the designs of the
respective objectives and demonstrate that TVM provides a favorable
quality--efficiency trade-off by achieving competitive image quality
with a single NFE.

\begin{figure}[t]
    \centering
    \includegraphics[width=\columnwidth]{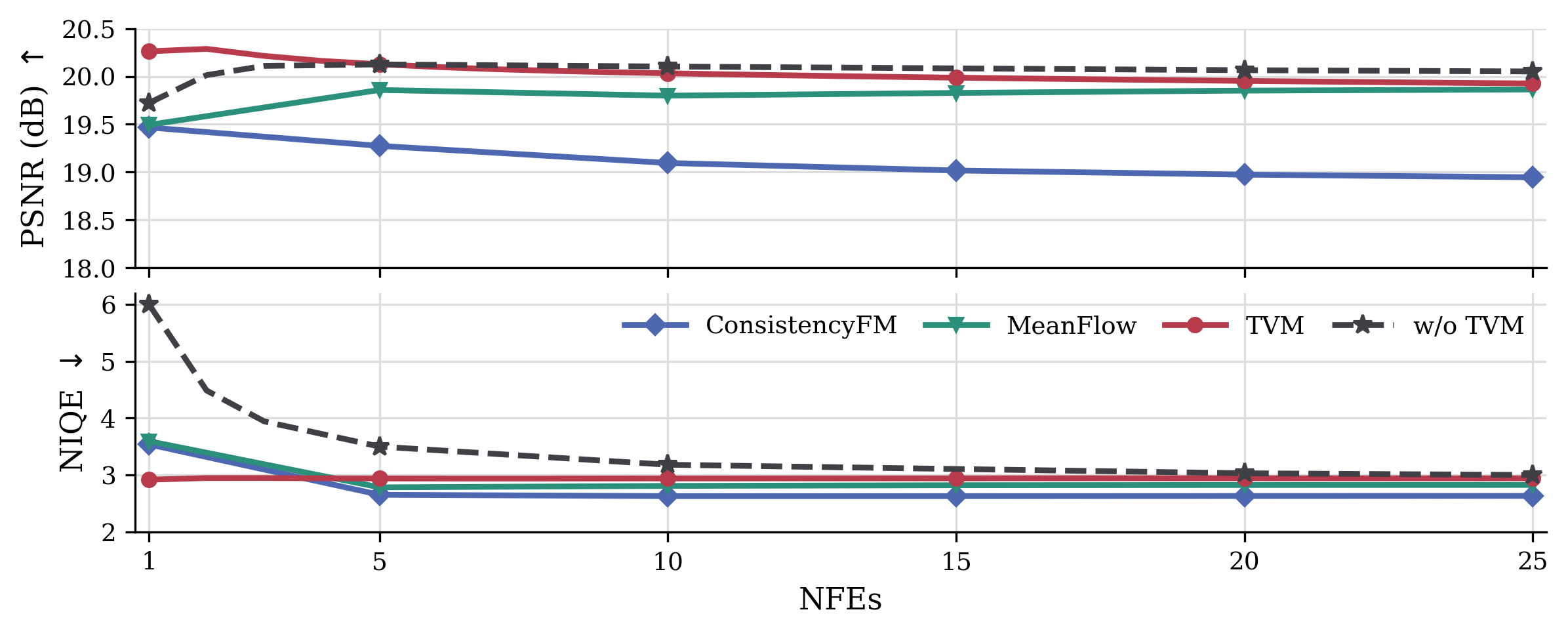}
    \caption{\textbf{One-Step Methods across NFEs.}
    PSNR and NIQE on LSD-DLL for TVM, MeanFlow, ConsistencyFM, and
    No-TVM as the number of NFEs varies.}
    \label{fig:nfe_objective_ablation}
\end{figure}

\paragraph{One-Step Methods across NFEs.}
\cref{fig:nfe_objective_ablation} extends the matched one-step
comparison by evaluating TVM, MeanFlow, ConsistencyFM, and No-TVM
across different inference budgets. At one NFE, TVM provides the
strongest joint distortion--perception performance. Its PSNR and NIQE
remain stable as the number of NFEs increases, indicating that it
already captures the enhancement trajectory in a single evaluation.
No-TVM improves substantially with additional NFEs but requires many
evaluations to approach one-step TVM. MeanFlow and ConsistencyFM improve
NIQE with additional NFEs but remain lower in PSNR, reflecting a
different distortion--perception trade-off. Overall, these results
further support TVM for efficient one-step LLIE.

\paragraph{Extended Ablation Study Table.}
\cref{tab:architecture_ablation_lsdu_supp} extends Tab.~5 from the main paper by adding CLIP-IQA~\cite{wang2023exploring} on LSD-DLL and reporting unseen-device generalization results on LSD-U.

\begin{table*}[t]
  \centering
  \scriptsize
  \setlength{\tabcolsep}{2.5pt}
  \resizebox{\textwidth}{!}{%
  \begin{tabular}{@{}lcccccccccccc@{}}
    \toprule
    \multirow{2}{*}{Variant} &
    \multirow{2}{*}{Latent} &
    \multirow{2}{*}{GFLOPs$\downarrow$} &
    \multirow{2}{*}{Params (M)$\downarrow$} &
    \multicolumn{6}{c}{LSD-DLL} &
    \multicolumn{3}{c}{LSD-U} \\
    \cmidrule(lr){5-10}\cmidrule(l){11-13}
    & & & &
    PSNR$\uparrow$ &
    SSIM$\uparrow$ &
    LPIPS$\downarrow$ &
    MUSIQ$\uparrow$ &
    NIQE$\downarrow$ &
    CLIP-IQA$\uparrow$ &
    NIQE$\downarrow$ &
    MUSIQ$\uparrow$ &
    CLIP-IQA$\uparrow$ \\
    \midrule

    Pixel Space (TVM) & None & 1693 & 121 &
    16.0 & 0.47 & 0.54 & 28.2 & 4.51 & 0.386 &
    6.15 & 25.2 & 0.210 \\

    SD-VAE (TVM) & SD-VAE & 471 & 205 &
    \mstdsci{19.6}{0.0287} &
    \mstdsci{0.504}{0.00342} &
    \mstdsci{0.472}{0.0326} &
    \mstdsci{\textbf{40.58}}{0.0984} &
    \mstdsci{3.372}{0.00104} &
    \mstdsci{0.609}{0.00077} &
    \mstdsci{4.35}{0.00120} &
    \mstdsci{\textbf{30.4}}{0.00574} &
    \mstdsci{0.362}{0.00016} \\

    No TVM (25 NFE) & TAESD & 415 & 124 &
    20.1 & 0.54 & \textbf{0.37} & 35.6 & 3.00 & \textbf{0.626} &
    4.75 & 28.8 & 0.341 \\

    ConsistencyFM & TAESD & 48 & 124 &
    19.46 & 0.52 & 0.50 & 32.8 & 3.53 & 0.573 &
    5.17 & 27.3 & 0.327 \\

    MeanFlow & TAESD & 48 & 124 &
    19.49 & 0.52 & 0.49 & 31.7 & 3.59 & 0.567 &
    5.32 & 27.8 & 0.344 \\

    \specialrule{0.3pt}{1pt}{1pt}

    \method{} & TAESD & 48 & 124 &
    \textbf{20.3} & \textbf{0.55} & 0.48 & 36.0 &
    \textbf{2.92} & 0.625 &
    \textbf{4.33} & 28.4 & \textbf{0.367} \\
    \bottomrule
  \end{tabular}}
  \caption{\textbf{Extended ablation study.}
  LSD-DLL and LSD-U results assessing the latent representation and
  flow-matching objective. SD-VAE results are reported as the mean
  $\pm$ sample standard deviation across five seeds. All other variants
  are deterministic and are therefore reported as single values.
  Best mean results are shown in \textbf{bold}.}
  \label{tab:architecture_ablation_lsdu_supp}
\end{table*}

\paragraph{Autoencoder Reconstruction Analysis.}
The extended architecture ablation in
\cref{tab:architecture_ablation_lsdu_supp} first evaluates the effect
of the latent autoencoder on end-to-end enhancement. Compared with the
SD-VAE-based variant, the TAESD-based \method{} achieves higher PSNR
and SSIM and lower NIQE while reducing the computational cost from
$471$ to $48$ GFLOPs and the parameter count from $205$M to $124$M.
The SD-VAE variant achieves lower LPIPS and higher MUSIQ, reflecting a
different distortion--perception trade-off.

To separate transport-model error from the reconstruction error
introduced by the frozen autoencoder, we additionally evaluate direct
target reconstruction,
$\tilde{x}_0=\mathcal{D}(\mathcal{E}(x_0))$, for TAESD and SD-VAE
across the three paired benchmarks. This setting represents the
performance obtained when the target encoding is recovered exactly and
therefore provides a useful reconstruction reference, but not a strict
bound on the output of \method{}. In particular, PSNR, SSIM, and LPIPS
measure correspondence with the paired target, whereas the no-reference
metrics NIQE and MUSIQ assess perceptual image statistics and may
improve even when reconstruction fidelity decreases. We therefore
report NIQE and MUSIQ only as complementary diagnostics.

\begin{table}[t]
    \centering
    \small
    \setlength{\tabcolsep}{3.5pt}
    \resizebox{\columnwidth}{!}{%
    \begin{tabular}{@{}llccccc@{}}
        \toprule
        Dataset &
        Setting &
        PSNR$\uparrow$ &
        SSIM$\uparrow$ &
        LPIPS$\downarrow$ &
        NIQE$\downarrow$ &
        MUSIQ$\uparrow$ \\
        \midrule

        \multirow{3}{*}{LOL-v1}
        & SD-VAE Rec.
        & 28.14 & 0.854 & 0.048 & 4.339 & 72.45 \\
        & TAESD Rec.
        & 30.33 & 0.909 & 0.031 & 3.735 & 72.70 \\
        \cmidrule(lr){2-7}
        & \method{}
        & 21.8 & 0.828 & 0.29 & 3.71 & 70.4 \\

        \midrule

        \multirow{3}{*}{LSD-DLL}
        & SD-VAE Rec.
        & 26.40 & 0.687 & 0.086 & 3.719 & 37.94 \\
        & TAESD Rec.
        & 29.49 & 0.816 & 0.069 & 3.350 & 35.77 \\
        \cmidrule(lr){2-7}
        & \method{}
        & 20.3 & 0.548 & 0.48 & 2.92 & 36.0 \\

        \midrule

        \multirow{3}{*}{UHD-LOL4K}
        & SD-VAE Rec.
        & 32.89 & 0.948 & 0.026 & 4.745 & 34.84 \\
        & TAESD Rec.
        & 36.11 & 0.981 & 0.019 & 4.552 & 34.48 \\
        \cmidrule(lr){2-7}
        & \method{}
        & 31.7 & 0.969 & 0.12 & 4.59 & 35.2 \\

        \bottomrule
    \end{tabular}}
    \caption{\textbf{Autoencoder reconstruction analysis.}
    Direct target reconstructions $\mathcal{D}(\mathcal{E}(x_0))$ are compared with the corresponding
    \method{}-XL enhancement outputs.}
    \label{tab:vae_reconstruction_reference}
\end{table}

\begin{table}[t]
    \centering
    \small
    \setlength{\tabcolsep}{6pt}
    \begin{tabular}{@{}lccc@{}}
        \toprule
        Autoencoder &
        LOL-v1 &
        LSD-DLL &
        UHD-LOL4K \\
        \midrule
        TAESD  & 1.181 & 1.792 & 0.266 \\
        SD-VAE & 1.963 & 4.579 & 0.693 \\
        \bottomrule
    \end{tabular}
    \caption{\textbf{Autoencoder reconstruction error.}
    MSE$\downarrow$ is reported in units of $\times 10^{-3}$.}
    \label{tab:vae_reconstruction_mse}
\end{table}

As shown in \cref{tab:vae_reconstruction_reference}, direct TAESD
reconstruction achieves higher PSNR and SSIM and lower LPIPS than
SD-VAE reconstruction on all three datasets. This differs from the
end-to-end comparison above, in which the SD-VAE variant achieves
slightly better LPIPS, because direct reconstruction isolates the
autoencoder from the transport model. On LSD-DLL, direct TAESD
reconstruction obtains $29.49$\,dB PSNR and $0.069$ LPIPS, compared
with $20.3$\,dB and $0.48$ for \method{}. This substantial gap
indicates considerable reconstruction headroom and suggests that
TAESD reconstruction error is not the principal bottleneck in
end-to-end enhancement performance.

\cref{tab:vae_reconstruction_mse} further reports the empirical
autoencoder reconstruction term, $\epsilon_{\mathrm{VAE}}$, measured
using MSE. TAESD obtains MSE values of $1.181$, $1.792$, and
$0.266$, compared with $1.963$, $4.579$, and $0.693$ for SD-VAE on
LOL-v1, LSD-DLL, and UHD-LOL4K, respectively (all $\times 10^{-3}$).
Thus, TAESD yields a smaller measurable autoencoder reconstruction term
on all three benchmarks. These values characterize only the
autoencoder component of the $W_2$ analysis from \cref{thm:pixel_w2_tvm_bound}, rather than the complete
bound, whose transport weighting and decoder-dependent constants are
not estimated.

Although TAESD uses a substantially smaller encoder and decoder,
parameter count alone does not determine the information retained by
the latent representation. At each latent spatial location, TAESD
encodes 16 channels, whereas SD-VAE encodes four. Because both use the
same spatial downsampling factor, TAESD produces four times as many
latent coefficients for a given input resolution. This higher latent
dimensionality helps explain its stronger reconstruction performance
despite its smaller model size.

\begin{figure}[t]
  \centering
  \includegraphics[width=\linewidth]{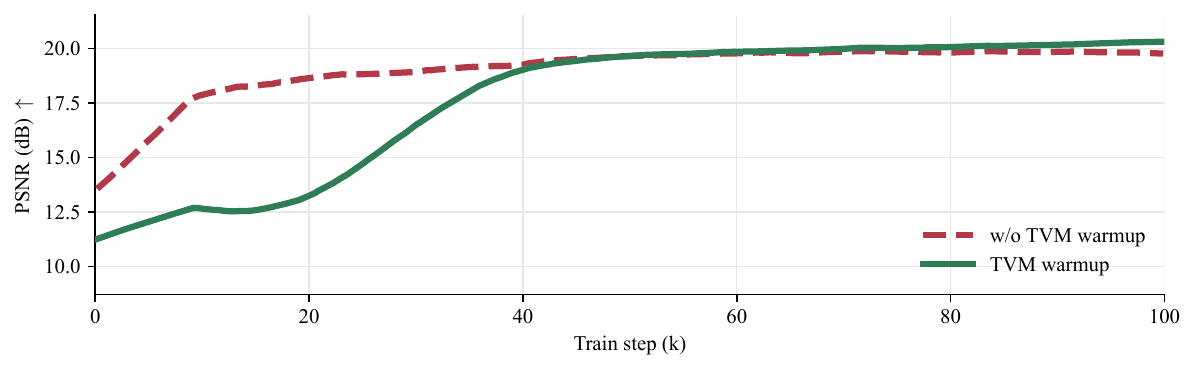}
  \vspace{2mm}
  \includegraphics[width=\linewidth]{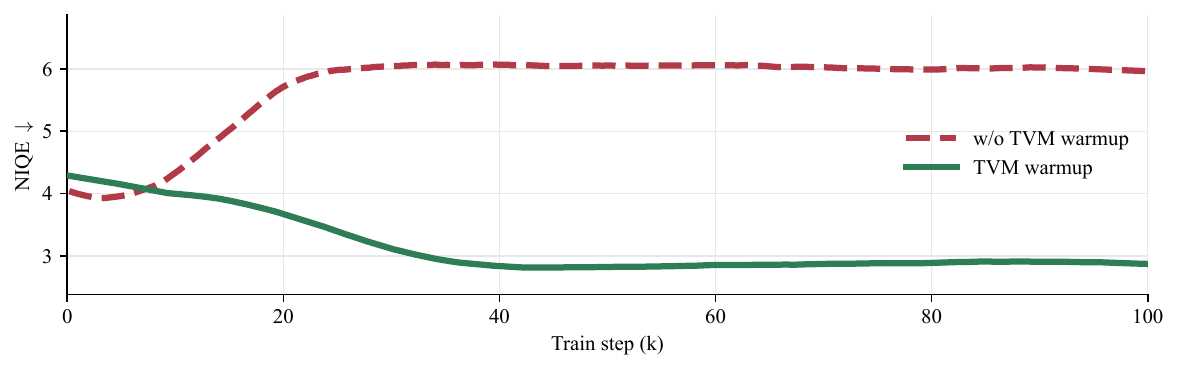}
  \caption{\textbf{TVM warmup ablation.} PSNR and NIQE on LSD-DLL with and without TVM warmup. Curves are smoothed for visualization.}
  \label{fig:tvm_warmup_ablation}
\end{figure}

\paragraph{TVM Warmup Ablation.}
\cref{fig:tvm_warmup_ablation} studies the effect of gradually introducing the TVM objective during training. We compare the proposed warmup-ramp schedule against a variant trained without TVM warmup.

Without TVM warmup, the TVM objective is introduced before the flow-matching model has learned a meaningful low-to-normal-light transport. As a result, the displacement objective is optimized with respect to an immature velocity field, leading the optimization toward an inferior solution. In contrast, the proposed warmup schedule allows the flow-matching objective to first establish a reliable transport before introducing the TVM objective, resulting in improved PSNR and substantially better NIQE at convergence.

\paragraph{Effect of source-latent perturbation.}
During training, we perturb the low-light source latent as
$\tilde{z}_1=z_1+\sigma\epsilon$, where
$\epsilon\sim\mathcal{N}(0,I)$. We evaluate four noise levels using distortion and perceptual metrics
on the LSD-DLL validation set, unseen-device generalization on LSD-U,
and cross-dataset mean NIQE on DICM, LIME, MEF, NPE, and VV, following
the evaluation in main-paper Tab.~4. As shown in \cref{tab:source_noise_ablation}, $\sigma=0.01$ improves
LSD-DLL PSNR and LPIPS over $\sigma=0$ and achieves the best
generalization results on LSD-U and across the five cross-dataset
benchmarks. Although $\sigma=0.10$ obtains better in-domain NIQE,
$\sigma=0.01$ provides the best overall balance when jointly
considering distortion, perceptual quality, and generalization.

\begin{table}[t]
  \centering
  \small
  \setlength{\tabcolsep}{3pt}
  \resizebox{\linewidth}{!}{%
  \begin{tabular}{@{}cccccc@{}}
    \toprule
    \multirow{2}{*}{Training $\sigma$} &
    \multicolumn{3}{c}{LSD-DLL} &
    LSD-U &
    Cross-dataset \\
    \cmidrule(lr){2-4}
    \cmidrule(lr){5-5}
    \cmidrule(l){6-6}
    & PSNR$\uparrow$
    & LPIPS$\downarrow$
    & NIQE$\downarrow$
    & NIQE$\downarrow$
    & Mean NIQE$\downarrow$ \\
    \midrule
    0.00 & 20.05 & 0.49 & 2.93 & 4.72 & 2.87 \\
    0.01 & \textbf{20.30} & \textbf{0.48} & 2.92
         & \textbf{4.33} & \textbf{2.86} \\
    0.05 & 19.91 & 0.50 & 2.90 & 4.88 & 2.90 \\
    0.10 & 19.62 & 0.51 & \textbf{2.84} & 4.67 & 2.98 \\
    \bottomrule
  \end{tabular}}
  \caption{\textbf{Effect of source-latent perturbation.}
  LSD-U measures unseen-device generalization, while cross-dataset
  mean NIQE is averaged over DICM, LIME, MEF, NPE, and VV.
  Best results are shown in \textbf{bold}.}
  \label{tab:source_noise_ablation}
\end{table}

\section{Additional Quantitative Results}
\label[appendix]{sec:supp_quantitative}

This section extends the quantitative comparisons presented in the main paper. Specifically, we report results over five evaluation seeds as $\text{mean}\pm\text{std}$ and include CLIP-IQA for all benchmarks.
For LOL-v1, we additionally report SSIM and LPIPS.
The extended results for LOL-v1, LSD-DLL, UHD-LOL4K, and LSD-U are reported in \cref{tab:lolv1_full,tab:lsd_dll_full,tab:uhd_lol4k_full,tab:lsd_u_full}, respectively.

\begin{table*}[t]
  \centering
  \small
  \setlength{\tabcolsep}{2.5pt}
  \resizebox{\textwidth}{!}{%
  \begin{tabular}{@{}lccccccccc@{}}
    \toprule
    Method &
    NFEs$\downarrow$ &
    GFLOPs$\downarrow$ &
    Params (M)$\downarrow$ &
    PSNR$\uparrow$ &
    SSIM$\uparrow$ &
    LPIPS$\downarrow$ &
    MUSIQ$\uparrow$ &
    NIQE$\downarrow$ &
    CLIP-IQA$\uparrow$ \\
    \midrule

    PyDiff~\cite{zhou2023pydiff}
    & 4 & 473 & 97.9
    & \mstdsci{\underline{21.9}}{0.2044}
    & \mstdsci{\textbf{0.895}}{0.0206}
    & \mstdsci{\textbf{0.11}}{0.0032}
    & \mstdsci{\textbf{74.2}}{0.1738}
    & \mstdsci{5.40}{0.1014}
    & \mstdsci{0.44}{0.0230} \\

    WCDM~\cite{wcdm}
    & 10 & 90 & 22.1
    & \mstdsci{\textbf{26.3}}{0.0118}
    & \mstdsci{\underline{0.887}}{0.0225}
    & \mstdsci{\underline{0.12}}{0.0006}
    & \mstdsci{66.9}{0.0104}
    & \mstdsci{4.52}{0.0112}
    & \mstdsci{0.40}{0.0003} \\

    ReDDiT~\cite{reddit}
    & 2 & 128 & 17.4
    & \mstdsci{20.8}{0.1545}
    & \mstdsci{0.883}{0.0022}
    & \mstdsci{0.20}{0.0022}
    & \mstdsci{\underline{71.7}}{0.0714}
    & \mstdsci{5.38}{0.0189}
    & \mstdsci{\textbf{0.70}}{0.0043} \\

    \midrule

    \textbf{\method{}-XL}
    & 1 & 48 & 124
    & 21.8 & 0.828 & 0.29 & 70.4
    & \underline{3.71} & \underline{0.554} \\

    \textbf{\method{}-M}
    & 1 & 36 & 21.8
    & 21.3 & 0.826 & 0.27
    & 71.5 & 3.79 & 0.547 \\

    \textbf{\method{}-S}
    & 1 & 34 & 9.1
    & 21.2 & 0.822 & 0.27 & 71.3
    & \textbf{3.70} & 0.549 \\

    \bottomrule
  \end{tabular}}
  \caption{\textbf{Five-seed evaluation results on LOL-v1.}}
  \label{tab:lolv1_full}
  \vspace{-10pt}
\end{table*}

\begin{table*}[t]
  \centering
  \small
  \setlength{\tabcolsep}{2.5pt}
  \resizebox{\textwidth}{!}{%
  \begin{tabular}{@{}lccccccccc@{}}
    \toprule
    Method &
    NFEs$\downarrow$ &
    GFLOPs$\downarrow$ &
    Params (M)$\downarrow$ &
    PSNR$\uparrow$ &
    SSIM$\uparrow$ &
    LPIPS$\downarrow$ &
    MUSIQ$\uparrow$ &
    NIQE$\downarrow$ &
    CLIP-IQA$\uparrow$ \\
    \midrule

    Retinexformer~\cite{cai2023retinexformer}
    & 1 & 17 & 1.6
    & 16.9
    & \textbf{0.567}
    & 0.54
    & 28.1
    & 4.64
    & 0.32 \\

    HVI-CIDNet~\cite{yan2025hvi}
    & 1 & 8 & 2.0
    & 15.2
    & 0.495
    & 0.56
    & 23.5
    & 4.42
    & 0.37 \\

    Multinex~\cite{multinex2026}
    & 1 & 3 & 0.04
    & 15.6
    & 0.501
    & 0.63
    & 21.6
    & 7.73
    & 0.39 \\

    TFFormer~\cite{lsd_dataset}
    & 1 & 34 & 5.9
    & 14.1
    & 0.349
    & 0.63
    & 32.7
    & 4.49
    & 0.45 \\

    \midrule
    
    PyDiff~\cite{zhou2023pydiff}
    & 4 & 473 & 97.9
    & \mstdsci{17.3}{0.0851}
    & \mstdsci{0.544}{0.0013}
    & \mstdsci{0.56}{0.0053}
    & \mstdsci{33.2}{0.1199}
    & \mstdsci{7.38}{0.0658}
    & \mstdsci{0.425}{0.0017} \\

    WCDM~\cite{wcdm}
    & 10 & 90 & 22.1
    & \mstdsci{17.1}{0.1346}
    & \mstdsci{\underline{0.553}}{0.0019}
    & \mstdsci{0.59}{0.0007}
    & \mstdsci{25.5}{0.0288}
    & \mstdsci{5.94}{0.0442}
    & \mstdsci{0.269}{0.0007} \\

    ReDDiT~\cite{reddit}
    & 2 & 128 & 17.4
    & \mstdsci{14.1}{0.0074}
    & \mstdsci{0.488}{0.0002}
    & \mstdsci{0.59}{0.0019}
    & \mstdsci{30.9}{0.1373}
    & \mstdsci{7.45}{0.0367}
    & \mstdsci{0.359}{0.0020} \\

    \midrule

    \textbf{\method{}-XL}
    & 1 & 48 & 124
    & \textbf{20.3}
    & 0.548
    & \textbf{0.48}
    & \underline{36.0}
    & 2.92
    & \textbf{0.625} \\

    \textbf{\method{}-M}
    & 1 & 36 & 21.8
    & 19.5
    & 0.531
    & \underline{0.50}
    & 35.5
    & \underline{2.86}
    & \underline{0.619} \\

    \textbf{\method{}-S}
    & 1 & 34 & 9.1
    & \underline{19.7}
    & 0.516
    & 0.52
    & \textbf{37.1}
    & \textbf{2.70}
    & 0.613 \\

    \bottomrule
  \end{tabular}}
  \caption{\textbf{Five-seed evaluation results on LSD-DLL.}}
  \label{tab:lsd_dll_full}
  \vspace{-10pt}
\end{table*}

\begin{table*}[t]
  \centering
  \small
  \setlength{\tabcolsep}{2.5pt}
  \resizebox{\textwidth}{!}{%
  \begin{tabular}{@{}lccccccccc@{}}
    \toprule
    Method &
    NFEs$\downarrow$ &
    GFLOPs$\downarrow$ &
    Params (M)$\downarrow$ &
    PSNR$\uparrow$ &
    SSIM$\uparrow$ &
    LPIPS$\downarrow$ &
    MUSIQ$\uparrow$ &
    NIQE$\downarrow$ &
    CLIP-IQA$\uparrow$ \\
    \midrule

    Retinexformer~\cite{cai2023retinexformer}
    & 1 & 17 & 1.6
    & 30.3
    & 0.983
    & 0.05
    & 34.0
    & \underline{4.47}
    & 0.286 \\

    HVI-CIDNet~\cite{yan2025hvi}
    & 1 & 8 & 2.0
    & \textbf{33.7}
    & \textbf{0.989}
    & \textbf{0.02}
    & 33.1
    & 5.03
    & 0.281 \\

    Multinex~\cite{multinex2026}
    & 1 & 3 & 0.04
    & 24.8
    & 0.958
    & 0.09
    & 32.0
    & 4.72
    & 0.293 \\

    TFFormer~\cite{lsd_dataset}
    & 1 & 34 & 5.9
    & 31.4
    & \underline{0.987}
    & \underline{0.04}
    & 32.9
    & 4.65
    & 0.267 \\

    \midrule

    PyDiff~\cite{zhou2023pydiff}
    & 4 & 473 & 97.9
    & \mstdsci{22.5}{0.3310}
    & \mstdsci{0.956}{0.0020}
    & \mstdsci{0.17}{0.0576}
    & \mstdsci{32.447}{0.7027}
    & \mstdsci{4.83}{0.0435}
    & \mstdsci{\textbf{0.331}}{0.0059} \\

    WCDM~\cite{wcdm}
    & 10 & 90 & 22.1
    & \mstdsci{26.8}{0.0469}
    & \mstdsci{0.957}{0.00002}
    & \mstdsci{0.16}{0.0312}
    & \mstdsci{31.728}{0.0964}
    & \mstdsci{\textbf{4.08}}{0.0388}
    & \mstdsci{0.231}{0.0097} \\

    ReDDiT~\cite{reddit}
    & 2 & 128 & 17.4
    & \mstdsci{22.2}{0.10582}
    & \mstdsci{0.924}{0.0036}
    & \mstdsci{0.25}{0.0033}
    & \mstdsci{\textbf{35.933}}{0.0209}
    & \mstdsci{5.24}{0.0035}
    & \mstdsci{0.301}{0.0002} \\

    \midrule

    \textbf{\method{}-XL}
    & 1 & 48 & 124
    & \underline{31.7}
    & 0.969
    & 0.12
    & \underline{35.2}
    & 4.59
    & \underline{0.323} \\

    \textbf{\method{}-M}
    & 1 & 36 & 21.8
    & 30.6
    & 0.961
    & 0.14
    & 35.1
    & 4.63
    & \underline{0.323} \\

    \textbf{\method{}-S}
    & 1 & 34 & 9.1
    & 30.3
    & 0.962
    & 0.14
    & 34.881
    & 4.62
    & 0.316 \\

    \bottomrule
  \end{tabular}}
  \caption{\textbf{Five-seed evaluation results on UHD-LOL4K.}}
  \label{tab:uhd_lol4k_full}
  \vspace{-10pt}
\end{table*}

\begin{table*}[t]
  \centering
  \small
  \setlength{\tabcolsep}{3pt}
  \resizebox{\linewidth}{!}{%
  \begin{tabular}{@{}lcccccc@{}}
    \toprule
    Method &
    NFEs$\downarrow$ &
    GFLOPs$\downarrow$ &
    Params (M)$\downarrow$ &
    NIQE$\downarrow$ &
    MUSIQ$\uparrow$ &
    CLIP-IQA$\uparrow$ \\
    \midrule

    Retinexformer~\cite{cai2023retinexformer}
    & 1 & 17 & 1.6
    & 7.35
    & 23.4
    & 0.205 \\

    HVI-CIDNet~\cite{yan2025hvi}
    & 1 & 8 & 2.0
    & 6.26
    & 26.5
    & 0.281 \\

    Multinex~\cite{multinex2026}
    & 1 & 3 & 0.04
    & 6.96
    & 27.2
    & 0.301 \\

    TFFormer~\cite{lsd_dataset}
    & 1 & 34 & 5.9
    & \underline{4.42}
    & 24.1
    & 0.271 \\

    \midrule
    
    PyDiff~\cite{zhou2023pydiff}
    & 4 & 473 & 97.9
    & \mstdsci{6.96}{0.0542}
    & \mstdsci{26.3}{0.1942}
    & \mstdsci{0.276}{0.0012} \\

    WCDM~\cite{wcdm}
    & 10 & 90 & 22.1
    & \mstdsci{6.16}{0.0241}
    & \mstdsci{23.692}{0.2839}
    & \mstdsci{0.183}{0.0034} \\

    ReDDiT~\cite{reddit}
    & 2 & 128 & 17.4
    & \mstdsci{7.94}{0.0188}
    & \mstdsci{26.371}{0.0501}
    & \mstdsci{0.265}{0.0012} \\

    \midrule

    \textbf{\method{}-XL}
    & 1 & 48 & 124
    & \textbf{4.33}
    & \textbf{28.4}
    & \textbf{0.367} \\

    \textbf{\method{}-M}
    & 1 & 36 & 21.8
    & 4.49
    & \underline{28.3}
    & \underline{0.358} \\

    \textbf{\method{}-S}
    & 1 & 34 & 9.1
    & 4.99
    & 27.7
    & 0.347 \\

    \bottomrule
  \end{tabular}}
  \caption{\textbf{Five-seed evaluation results on LSD-U.}}
  \label{tab:lsd_u_full}
  \vspace{-10pt}
\end{table*}

\paragraph{Runtime and Memory Analysis.}
\cref{tab:latency_memory} reports end-to-end inference latency and peak memory usage. All measurements use batch size one and report latency averaged over 100 timed inferences following 20 warm-up iterations. GPU measurements were obtained on a single NVIDIA H100 at resolutions from $256{\times}256$ to $4096{\times}4096$. We additionally evaluate $1024{\times}1024$ inference on an Apple M4 Max using ONNX with Arm SME/SME2 acceleration. SME is an Armv9-A instruction-set extension designed to accelerate matrix and vector processing~\cite{arm_sme_2024}.

On the H100, \method{}-S achieves the lowest latency and peak allocated memory across all evaluated resolutions. At high resolutions, all \method{} variants are substantially faster and require less GPU memory than the compared generative methods. Under Arm SME/SME2 execution, all \method{} variants also achieve lower latency than the competing
methods. These results demonstrate practical runtime efficiency across both server-class GPU acceleration and local, edge-oriented Arm execution.

\begin{table*}[t]
  \centering
  \small
  \setlength{\tabcolsep}{2.5pt}
  \resizebox{\textwidth}{!}{%
  \begin{tabular}{@{}lc*{12}{r}@{}}
    \toprule
    \multirow{3}{*}{Method} &
    \multirow{3}{*}{NFE} &
    \multicolumn{10}{c}{NVIDIA H100 GPU} &
    \multicolumn{2}{c}{Apple M4 Max (SME/SME2)} \\
    \cmidrule(lr){3-12}
    \cmidrule(l){13-14}
    & &
    \multicolumn{2}{c}{$256{\times}256$} &
    \multicolumn{2}{c}{$512{\times}512$} &
    \multicolumn{2}{c}{$1024{\times}1024$} &
    \multicolumn{2}{c}{$2048{\times}2048$} &
    \multicolumn{2}{c}{$4096{\times}4096$} &
    \multicolumn{2}{c}{$1024{\times}1024$} \\
    \cmidrule(lr){3-4}
    \cmidrule(lr){5-6}
    \cmidrule(lr){7-8}
    \cmidrule(lr){9-10}
    \cmidrule(lr){11-12}
    \cmidrule(l){13-14}
    & &
    Lat. (ms) & Mem. (MB) &
    Lat. (ms) & Mem. (MB) &
    Lat. (ms) & Mem. (MB) &
    Lat. (ms) & Mem. (MB) &
    Lat. (ms) & Mem. (MB) &
    Lat. (ms) & Mem. (MB) \\
    \midrule

    ReDDiT
    & 2
    & \underline{31.4} & 291.3
    & 65.7 & 859.1
    & 286.8 & 3130.4
    & 1791.4 & 12215.2
    & 39704.3 & 48554.7
    & 7484.1 & 25000.2 \\

    PyDiff
    & 4
    & 80.7 & 1099.9
    & 191.7 & 1949.3
    & 738.6 & 5346.7
    & 4485.5 & 18936.2
    & 80846.0 & 73294.4
    & 23682.8 & 27056.0 \\

    WCDM
    & 10
    & 166.8 & 195.4
    & 172.6 & 430.1
    & 229.8 & 1369.1
    & 974.7 & 5125.1
    & 7606.8 & 35277.1
    & 4987.1 & 4047.0 \\

    \midrule

    \textbf{\method{}-XL}
    & 1
    & 40.9 & 559.3
    & 45.8 & 708.3
    & 65.0 & 1304.6
    & 237.4 & 3689.6
    & 1149.1 & 13229.6
    & 3710.3 & 4557.8 \\

    \textbf{\method{}-M}
    & 1
    & 31.5 & \underline{165.5}
    & \underline{37.1} & \underline{314.5}
    & \underline{59.8} & \underline{910.8}
    & \underline{197.8} & \underline{3295.8}
    & \underline{870.2} & \underline{12835.8}
    & \underline{2709.6} & \underline{3526.0} \\

    \textbf{\method{}-S}
    & 1
    & \textbf{27.6} & \textbf{116.4}
    & \textbf{32.7} & \textbf{265.4}
    & \textbf{53.8} & \textbf{861.7}
    & \textbf{192.3} & \textbf{3246.7}
    & \textbf{773.5} & \textbf{12786.7}
    & \textbf{1811.2} & \textbf{3367.2} \\

    \bottomrule
  \end{tabular}}
  \caption{\textbf{Runtime efficiency on an NVIDIA H100 GPU and Arm SME/SME2.} We report mean end-to-end latency and peak memory usage with batch size one after 20 warm-up iterations and over 100 timed
  inferences. GPU results cover five input resolutions and use PyTorch peak allocated memory. SME/SME2 results are measured at $1024{\times}1024$ on an Apple M4 Max using ONNX, with memory measured as the peak resident-set size (RSS) of the inference process. Best results are shown in bold and second-best results are underlined.}
  \label{tab:latency_memory}
\end{table*}

\section{Additional Qualitative Results}
\label[appendix]{sec:supp_qualitative}

\begin{figure*}[p]
  \centering
  \includegraphics[width=\textwidth]{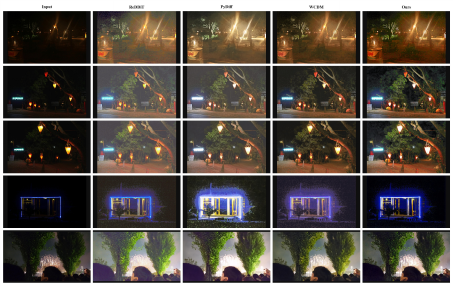}
  \caption{\textbf{Qualitative results on LSD-U.}}
  \label{fig:appendix_dll_unpair_qualitative}
\end{figure*}

\begin{figure*}[t]
  \centering
  \includegraphics[width=\textwidth]{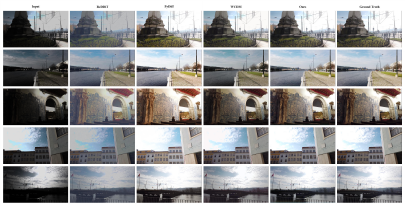}
  \caption{\textbf{Qualitative results on UHD-LOL4K.}}
  \label{fig:uhd_lol4k_qualitative_appendix}
\end{figure*}

\begin{figure*}[t]
  \centering
  \includegraphics[width=\textwidth]{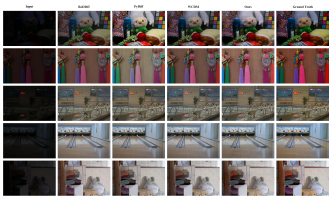}
  \caption{\textbf{Qualitative results on LOL-v1.}}
  \label{fig:lolv1_qualitative_appendix}
\end{figure*}

\begin{figure*}[p]
  \centering
  \includegraphics[width=\textwidth]{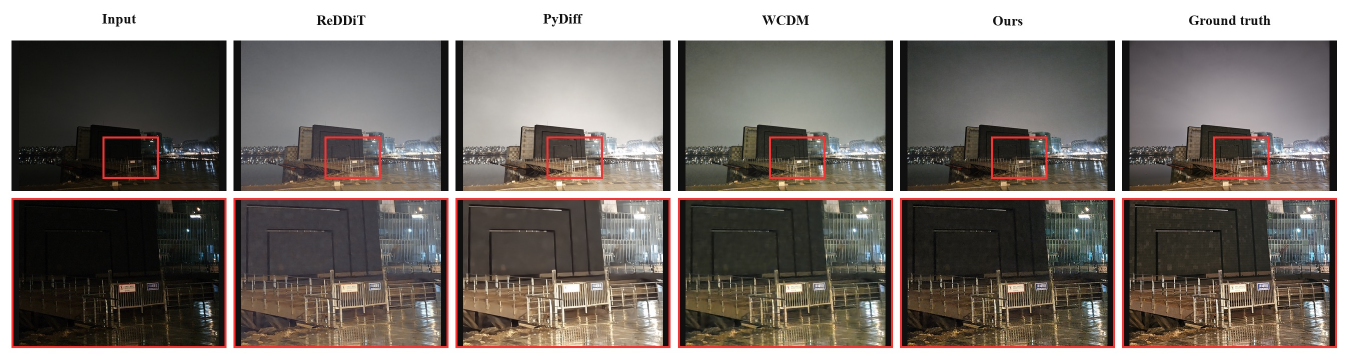}\\[-1pt]
  \includegraphics[width=\textwidth]{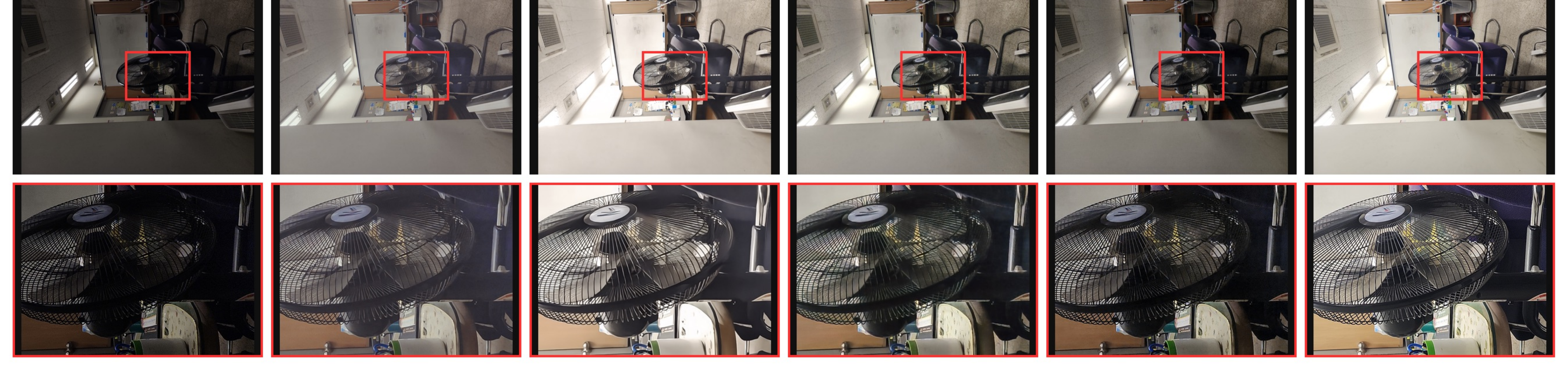}\\[-1pt]
  \includegraphics[width=\textwidth]{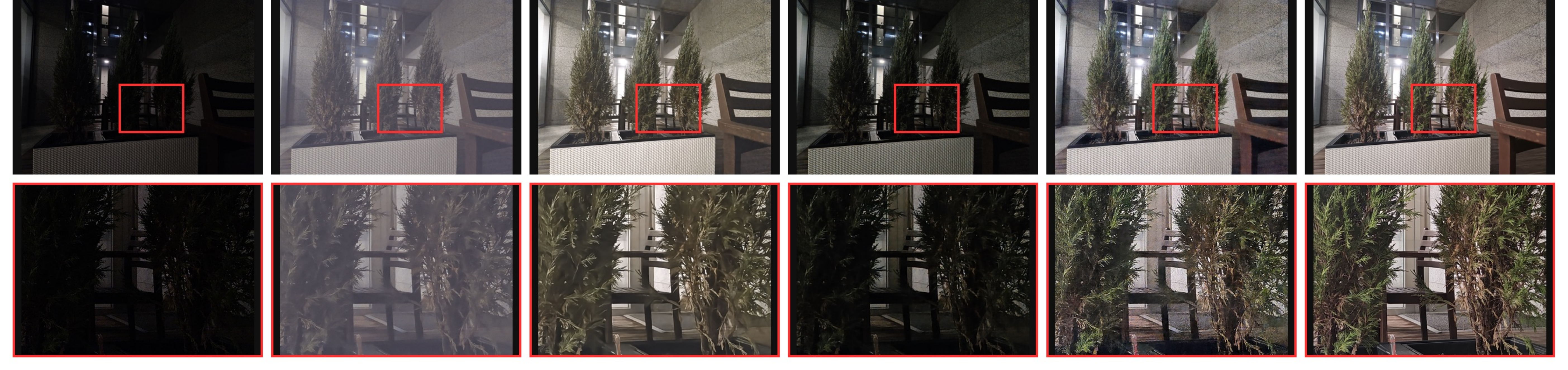}\\[-1pt]
  \includegraphics[width=\textwidth]{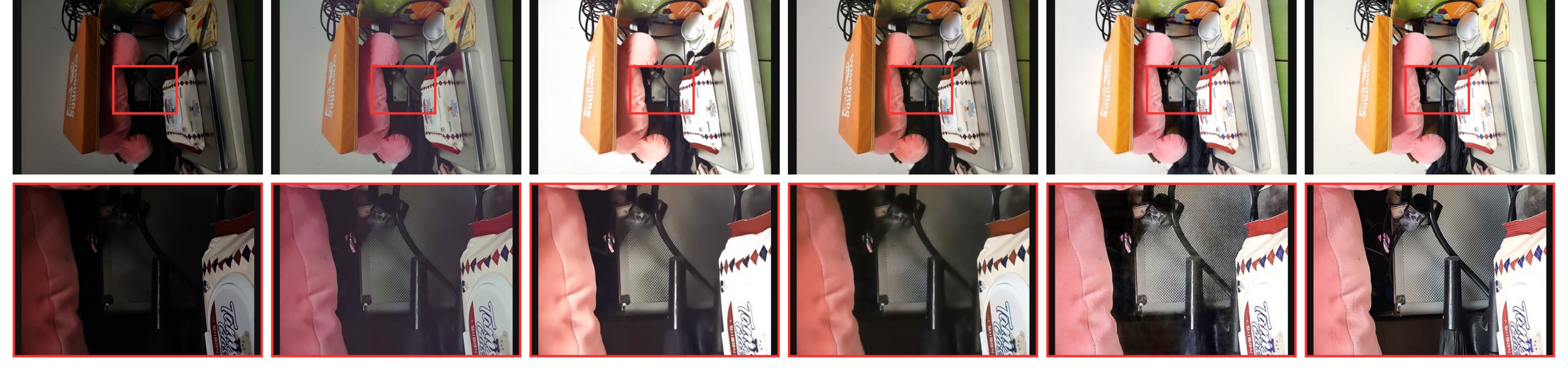}
  \caption[]{\textbf{Qualitative results on LSD-DLL.} Each full-image comparison row is followed by a magnified crop from the full-image.}
  \label{fig:more_lsd_dll_qualitative}
\end{figure*}

\begin{figure*}[p]
  \centering
  \includegraphics[width=\textwidth]{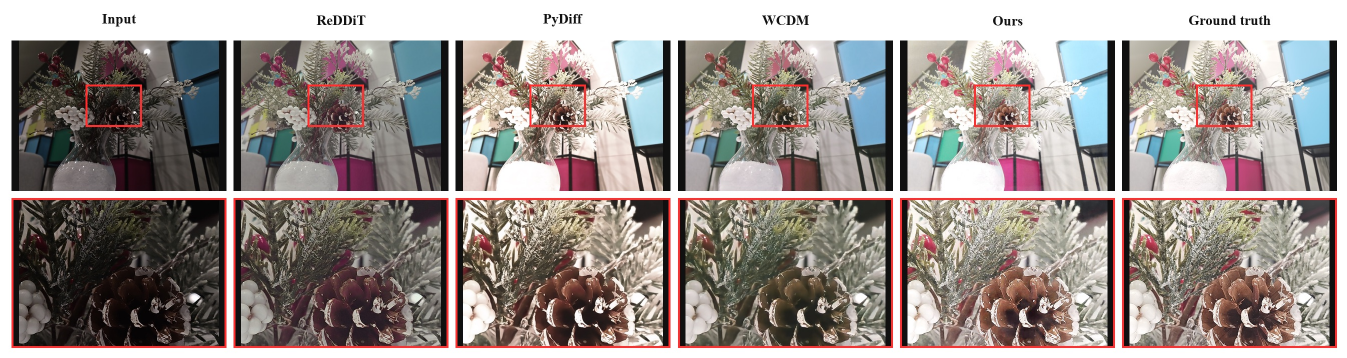}\\[-1pt]
  \includegraphics[width=\textwidth]{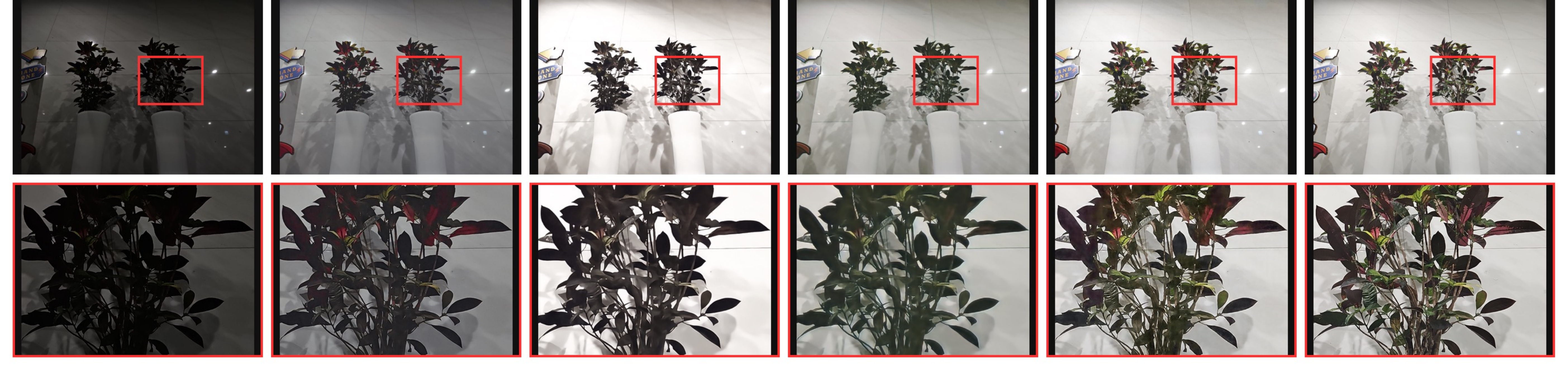}\\[-1pt]
  \includegraphics[width=\textwidth]{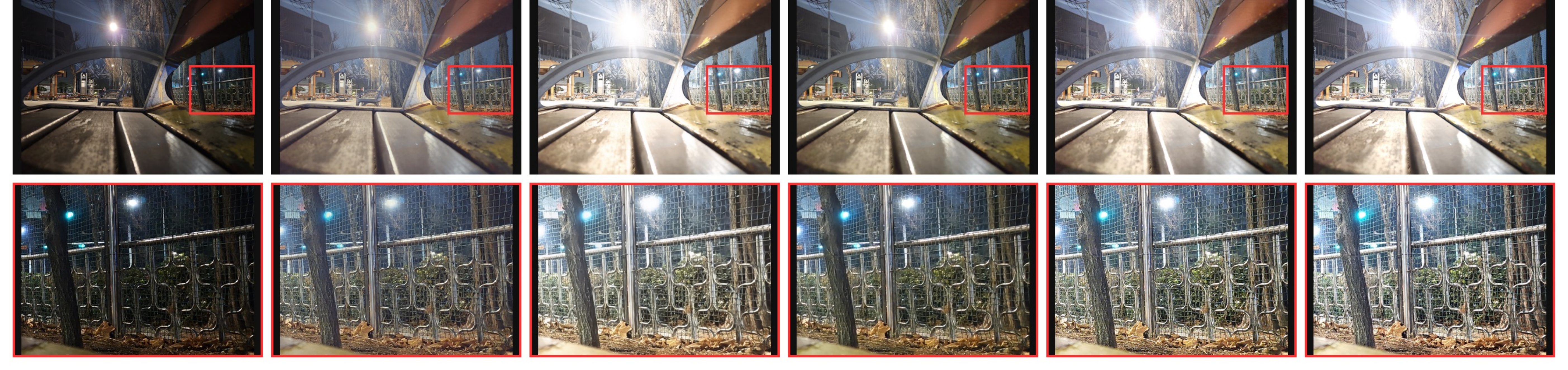}\\[-1pt]
  \includegraphics[width=\textwidth]{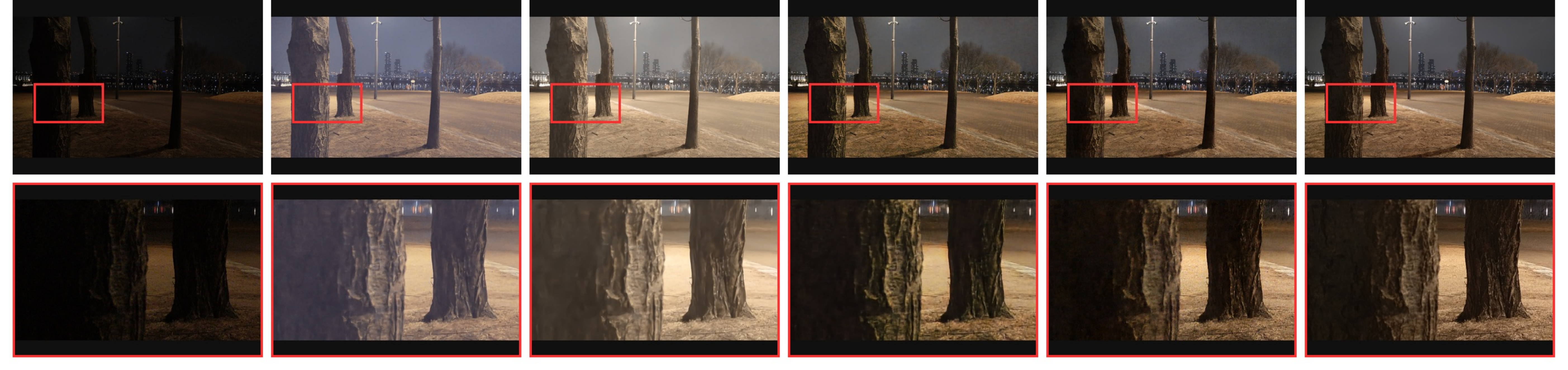}
  \caption[]{\textbf{Qualitative results on LSD-DLL.} Each full-image comparison row is followed by a magnified crop from the full-image.}
  \label{fig:more_lsd_dll_qualitative_2}
\end{figure*}

\begin{figure*}[p]
  \centering
  \includegraphics[width=\textwidth,height=0.93\textheight,keepaspectratio]{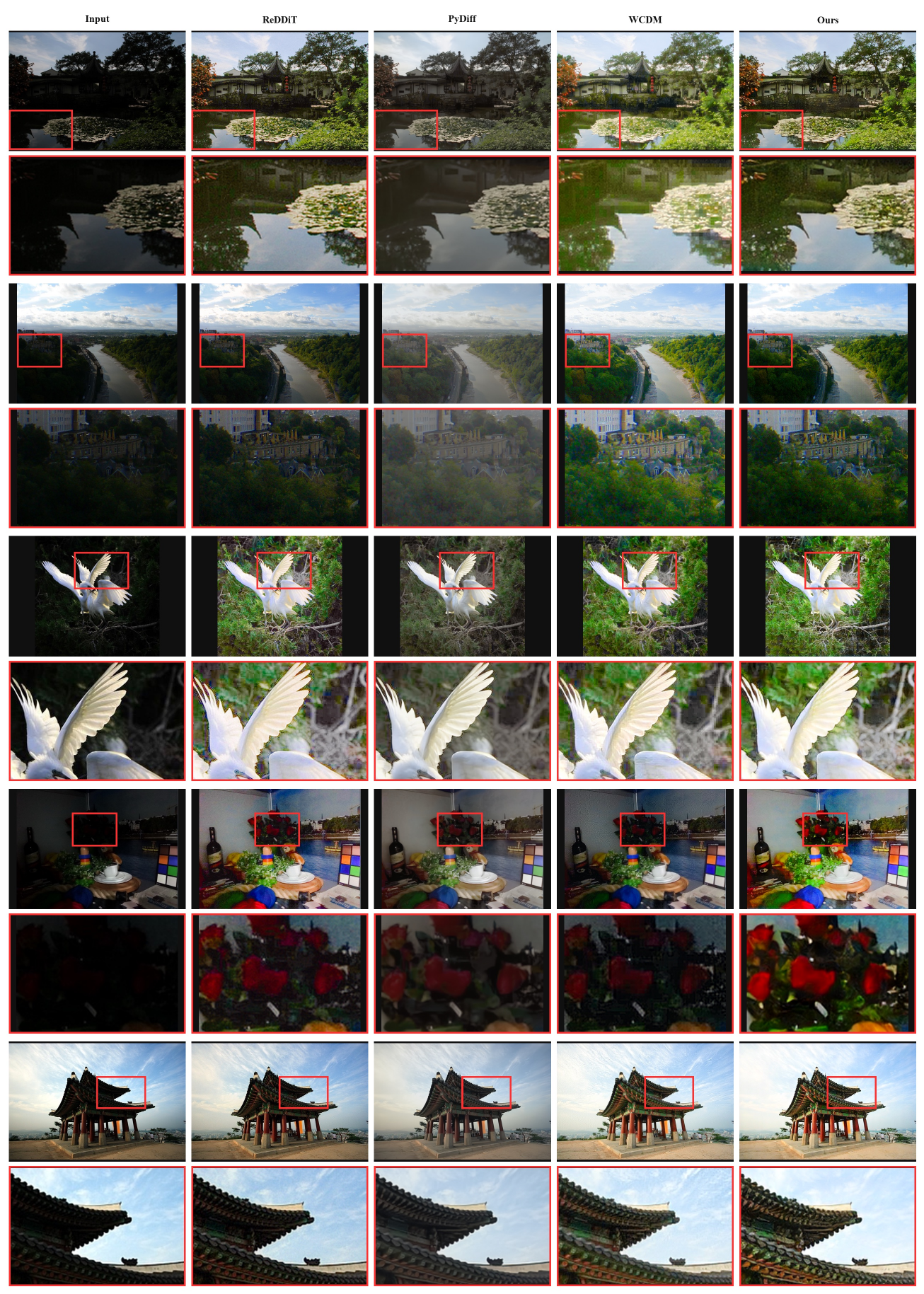}
  \caption{\textbf{Qualitative comparisons on the unpaired LIME, DICM, MEF, NPE, and VV datasets.} Each full-image comparison row is followed by a magnified crop from the full-image.}
  \label{fig:five_dataset_qualitative_appendix}
\end{figure*}

This section presents additional qualitative comparisons across several benchmarks, including unseen-device generalization on LSD-U (\cref{fig:appendix_dll_unpair_qualitative}), high-resolution enhancement on UHD-LOL4K (\cref{fig:uhd_lol4k_qualitative_appendix}) and LSD-DLL (\cref{fig:more_lsd_dll_qualitative}, \cref{fig:more_lsd_dll_qualitative_2}), and low-resolution enhancement on LOL-v1 (\cref{fig:lolv1_qualitative_appendix}). In \cref{fig:five_dataset_qualitative_appendix}, we further extend the qualitative generalization comparison shown in Fig.~4 of the main paper with magnified crops to enable closer inspection of dark regions and facilitate comparison with competing methods. Overall, the results show that \method{} produces natural illumination, preserves scene structure and fine details, and avoids severe visual artifacts across diverse low-light inputs.

\clearpage

\end{document}